\documentclass{article}

\usepackage{microtype}
\usepackage{graphicx}
\usepackage{subfigure}
\usepackage{booktabs} 

\usepackage{hyperref}

\usepackage[accepted]{icml2023}

\usepackage{amsmath}
\usepackage{amssymb}
\usepackage{mathtools}
\usepackage{amsthm}

\usepackage[capitalize,noabbrev]{cleveref}

\usepackage{verbatim}
\usepackage{graphicx}
\usepackage{booktabs} 
\usepackage{caption}
\usepackage{wrapfig}
\usepackage{multirow}

\theoremstyle{plain}

\theoremstyle{definition}

\theoremstyle{remark}

\usepackage[textsize=tiny]{todonotes}

\icmltitlerunning{Unsupervised Object-Centric Representations for Reinforcement Learning}

\begin{document}

\twocolumn[
\icmltitle{An Investigation into Pre-Training Object-Centric Representations\\for Reinforcement Learning}



\icmlsetsymbol{equal}{*}

\begin{icmlauthorlist}
\icmlauthor{Jaesik Yoon}{sap}
\icmlauthor{Yi-Fu Wu}{rutgers}
\icmlauthor{Heechul Bae}{etri}
\icmlauthor{Sungjin Ahn}{kaist}
\end{icmlauthorlist}

\icmlaffiliation{sap}{SAP}
\icmlaffiliation{rutgers}{Rutgers University}
\icmlaffiliation{etri}{ETRI}
\icmlaffiliation{kaist}{KAIST}

\icmlcorrespondingauthor{Jaesik Yoon and Sungjin Ahn}{mail@jaesikyoon.com and sjn.ahn@gmail.com}
\icmlkeywords{Machine Learning, ICML}

\vskip 0.3in
]



\printAffiliationsAndNotice{} 

\begin{abstract}
Unsupervised object-centric representation (OCR) learning has recently drawn attention as a new paradigm of visual representation. This is because of its \textit{potential} of being an effective pre-training technique for various downstream tasks in terms of sample efficiency, systematic generalization, and reasoning. Although image-based reinforcement learning (RL) is one of the most important and thus frequently mentioned such downstream tasks, the benefit in RL has surprisingly not been investigated systematically thus far. Instead, most of the evaluations have focused on rather indirect metrics such as segmentation quality and object property prediction accuracy. In this paper, we investigate the effectiveness of OCR pre-training for image-based reinforcement learning via empirical experiments. 
For systematic evaluation, we introduce a simple object-centric visual RL benchmark and conduct experiments to answer questions 
such as ``Does OCR pre-training improve performance on object-centric tasks?'' and ``Can OCR pre-training help with out-of-distribution generalization?''.
Our results provide empirical evidence for valuable insights into the effectiveness of OCR pre-training for RL and the potential limitations of its use in certain scenarios.
Additionally, this study also examines the critical aspects of incorporating OCR pre-training in RL, including performance in a visually complex environment and the appropriate pooling layer to aggregate the object representations. The benchmark and source code are available on the project website:\href{hhttps://sites.google.com/view/ocrl/home}{https://sites.google.com/view/ocrl/home}.
\vspace{-3mm}
\end{abstract}

\section{Introduction}

\begin{figure}[t]
  \begin{center}
   \includegraphics[width=0.98\linewidth]{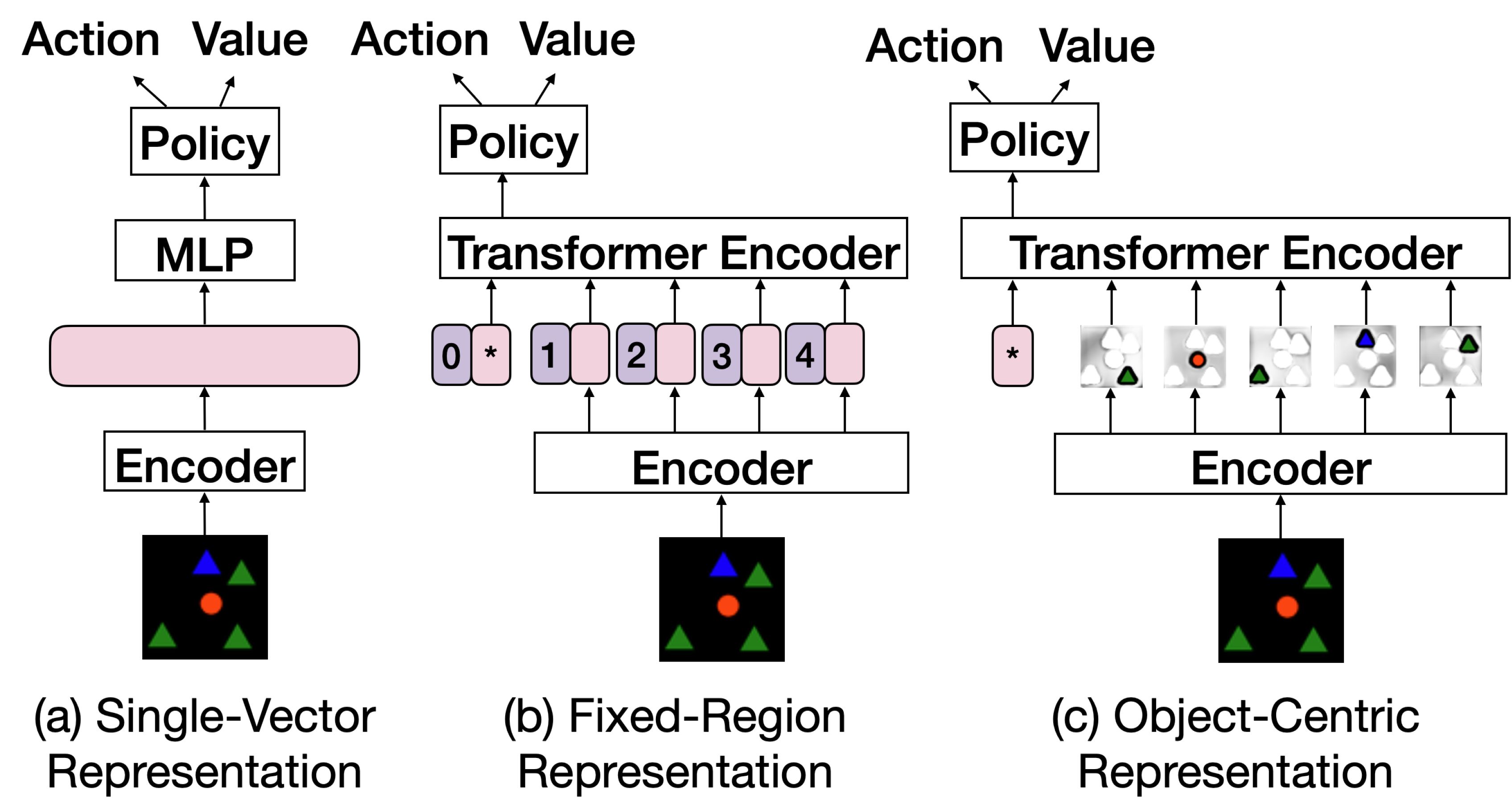}
  \end{center}
    \vspace{-3mm}
  \caption{The model architectures for representation types.}
  \label{fig:overall_archi}
    \vspace{-6mm}
\end{figure}

One of the main challenges in deep reinforcement learning (RL) from pixels is determining how to effectively represent the state of the environment.
Many previous approaches have represented the state using single-vector representations \citep{dqn, vae_ood,dqn,visuomotor}, encoding the entire input image into a single vector which is then used as input for the policy network (Figure \ref{fig:overall_archi}a). However, such representations may fail to capture important relationships and interactions between entities in the scene \citep{rn}. One way to address this limitation is to use a region-based representation, where an image is encoded into a grid of representations \citep{rdrl} which are then combined using a transformer encoder that allows for explicit modeling of interactions between the different regions (Figure \ref{fig:overall_archi}b).

Recent work in learning unsupervised Object-Centric Representations (OCR) provides another potentially promising way of representing the state of the scene \citep{air,spair,space,cswm, op3, monet,iodine,genesis,genesisv2,sa,slate}.
These approaches learn a structured visual representation from images without the need for labels, modeling each image as a composition of objects.
They provide a more semantically explicit way of modeling the entities in the scene than region-based representation models and can be similarly combined using a transformer encoder to model the relationships between the objects (Figure \ref{fig:overall_archi}c).
Thus, they have the promise of being an effective pre-training technique to learn representations for downstream RL.

Most previous research in OCRs, however, have evaluated OCRs only indirectly in the context of reconstruction loss, segmentation quality, or object property prediction \citep{generalization_ocr}. 
While some studies have applied OCRs to a few specific RL problems \citep{rim, smorl, cobra, load}, OCR pre-training has not been systematically and thoroughly evaluated for RL tasks. Therefore, many aspects of the relationship between OCR pre-training and RL remain unclear and further research is needed to fully understand its potential impact. This is of particular importance because many properties often show quite different behavior when it is applied to reinforcement learning due to the difficulties related to non-stationarity and reward sparsity.

\begin{figure*}[t]
    \centering
    \includegraphics[width=0.85\textwidth]{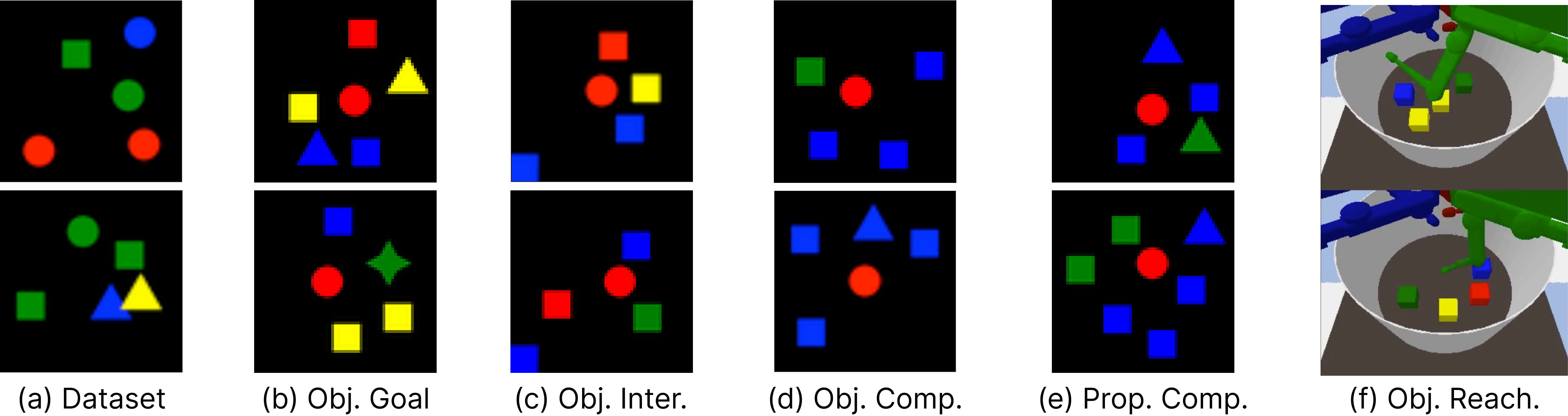}
    \caption{Samples from the dataset and five tasks in our benchmark; Object Goal / Object Interaction / Object Comparison / Property Comparison / Object Reaching tasks. In the 2D tasks ((b) - (e)), the red ball is always the agent. See main text for details about each task. In the robotics task (f), the goal is to use the green robotic finger to touch the blue object before touching any of the distractor objects.}
    \label{fig:task_illustration}
\end{figure*}

In this study, we investigate the effectiveness of OCR pre-training as a representation learning framework for RL through empirical experimentation. 
To achieve this, we propose a new benchmark that covers a range of object-centric tasks such as object interaction and relational reasoning \citep{stephens2008one}. The benchmark is set in a visually simple 2D scene (shown in Figure \ref{fig:task_illustration}) that current unsupervised OCR methods can successfully decompose, allowing us to isolate the effects of different experiment parameters and draw specific conclusions about when OCR pre-training is effective.
We further evaluate the performance in a more visually complex 3D scene \citep{causalworld} to explore a more realistic scenario where OCR pre-training may be beneficial.

\begin{figure}[t]
  \begin{center}
   \includegraphics[width=0.93\linewidth]{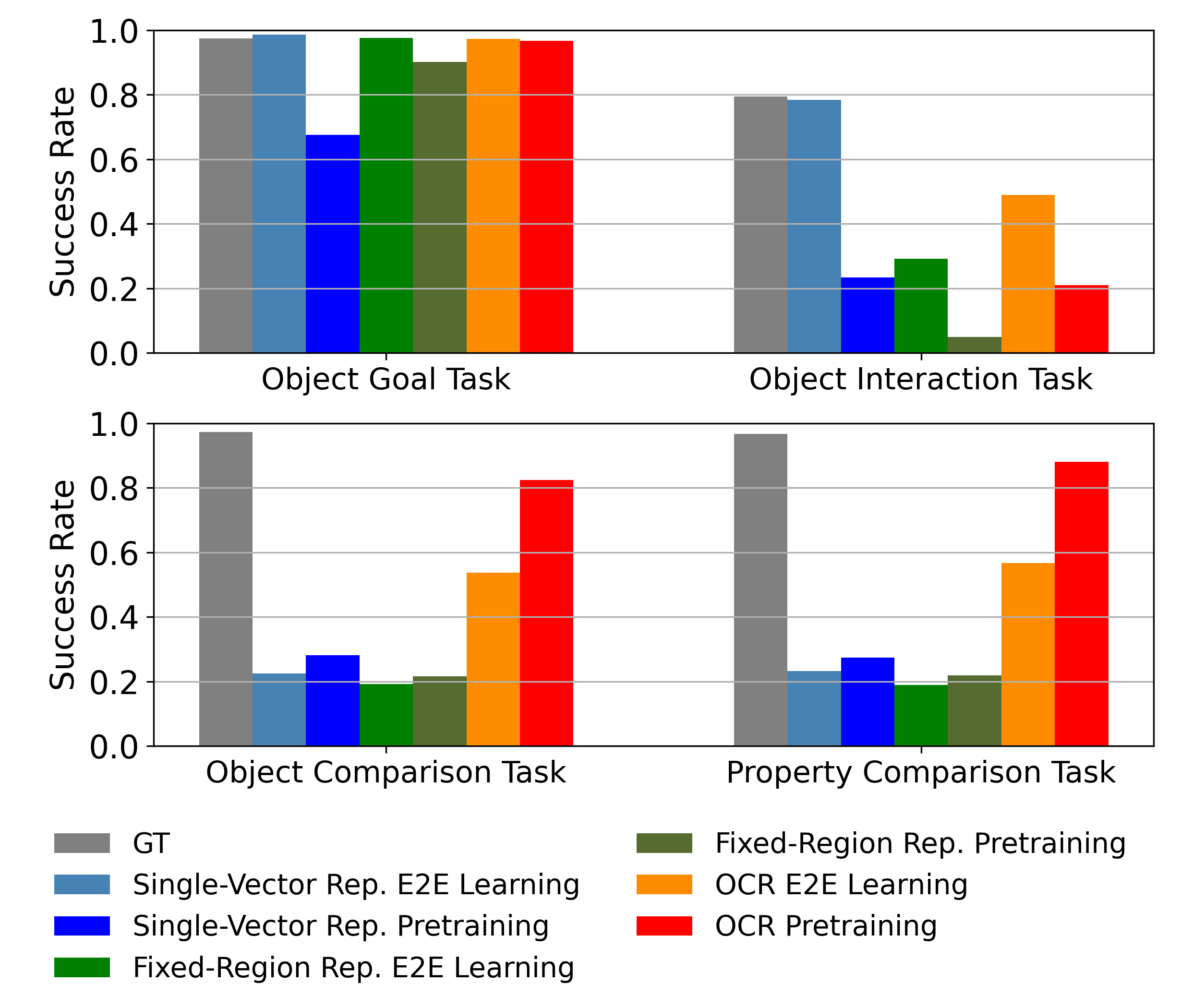}
    \vspace{-3mm}
  \end{center}
  \caption{The performance comparison of unsupervised object-centric representation (OCR) pre-training against other representation types in object-centric tasks. The results indicate that OCR pre-training demonstrates a significant performance gap compared to other representations and slightly worse performance than ground truth states in the comparison tasks where relational reasoning is a crucial aspect. However OCR pre-training performance is similar to or worse than baselines for other tasks.}
  \label{fig:overall_regime}
    \vspace{-2mm}
\end{figure}

Our investigation includes evaluating a series of hypotheses that have been presumed in prior work but not systematically investigated for OCR pre-training \citep{van2019perspective,building,binding, oomdp, schemanet, rdrl,lrn,rim,load,smorl}. The results of our investigation provide insights beyond these hypotheses. For instance, one of the hypotheses posits that OCR leads to improved performance in object-centric tasks \citep{smorl}.
However, our findings, as illustrated in Figure \ref{fig:overall_regime}, indicate that while OCR pre-training can indeed lead to improved performance in relational-reasoning tasks, it may not be as beneficial in other tasks even if they are object-centric.
Another common hypothesis suggests that decompositional representations are beneficial for reasoning tasks \citep{binding,van2019perspective,building}.
From our investigation, we find that the \textit{type} of the decompositional representation is crucial -- OCR pre-training performs well on relational-reasoning tasks, but fixed region representations, another type of decompositional representation, failed to solve these tasks.
Additionally, we also investigate important characteristics of applying OCR pre-training to RL, such as performance in a visually complex environment and what kind of pooling layer is appropriate to aggregate the object representations.

The main contribution of this paper is to provide insights into the effectiveness of OCR pre-training in RL tasks and the potential and limitations of its use through empirical evidence. To accomplish this, we make the following specific contributions: (1) propose a new simple benchmark to systematically validate OCR pre-training for RL tasks, (2) evaluate OCR pre-training performance compared with various baselines on this benchmark, and (3) systematically analyze different aspects of OCR pre-training to develop a better understanding of when and why OCR pre-training is beneficial for RL. Additionally, we release the benchmark and our experiment framework code to the community through the project website: \href{hhttps://sites.google.com/view/ocrl/home}{https://sites.google.com/view/ocrl/home}.

\section{Related Work} \label{sec:related_work}

\textbf{Object-Centric Representation Learning}.
Recent research in machine learning has focused on developing unsupervised object-centric representation (OCR) learning methods \citep{air,spair,sqair,space,scalor,cswm, root, swb, op3, monet,iodine,genesis,genesisv2,ocvt,sa,gswm,slate,savi,savi++,steve,sysbinder,ocsd}. These methods aim to learn structured visual representations from images without the need for labels, modeling each image as a composition of objects. The motivation behind this research is the potential benefits for downstream tasks such as improved generalization and relational reasoning \citep{binding, van2019perspective}.

In this paper, we investigate state-of-the-art OCR learning methods such as IODINE \citep{iodine}, Slot-Attention \citep{sa}, and SLATE \citep{slate}. IODINE represents objects with multiple latent variables through iterative refinement. Slot-Attention uses a similar approach but incorporates an attention mechanism to refine the variables. SLATE improves upon Slot-Attention by using a Transformer-based decoder instead of a pixel-mixture decoder, resulting in better generalization and comparable reconstruction performance. Additionally, we also implement a larger version of Slot-Attention (Slot-Attention-Large) to fairly compare the two methods with similar model sizes.

\textbf{Object-Centric Representations and Reinforcement Learning}. 
Reinforcement learning (RL) is a frequently mentioned downstream task where OCR are thought to be beneficial due to their potential for improved generalization, reasoning, and sample efficiency \citep{rdrl,garnelo2016towards,oomdp,schemanet,crafter_cnnfm,lrn,visuomotor}. However, to our knowledge, there have been no studies that systematically and thoroughly demonstrate these benefits.
\citet{rim} evaluated OCR for RL through end-to-end learning, which may lead to task-specific representations that lack the strengths of unsupervised OCR learning such as sample efficiency, generalization, and reasoning.
\citet{smorl} investigated OCR pre-training but applied a bounding box-based method \citep{scalor} and proposed/evaluated a new policy for the limited regime of goal-conditioned RL. 
\citet{cobra} evaluate OCR pre-training for a synthetic benchmark but a simple search is used rather than policy learning.

Previous studies have also investigated the use of decomposed representations in RL \citep{ke2021systematic,rdrl,garnelo2016towards,oomdp,schemanet,crafter_cnnfm,lrn,visuomotor}. Many of these works use CNN feature maps \citep{ke2021systematic,rdrl, crafter_cnnfm, visuomotor} as the representation or their own encoders \citep{garnelo2016towards}. Other studies have used ground truth states \citep{oomdp, schemanet, lrn}, and these representations have been implemented through separate object detectors and encoders \citep{oomdp, load}. The effectiveness of pre-trained representations for out-of-distribution generalization of RL agents is studied in \citep{vae_ood}, but only the single-vector representation (i.e. VAE) is evaluated. Causal discovery is explored by proposing a new benchmark in \citep{ke2021systematic}, but the evaluation focuses only on CNN feature maps and causality, excluding several hypotheses related to OCRs. In our work, we use a similar model as a baseline to compare with our pre-trained OCR model.

\section{Experimental Setup}

In this section, we provide an overview of our experimental setup.
We first discuss the OCR pre-training models and baselines we chose to evaluate and explain how the representations are used in a policy for RL.
We then introduce the tasks used in our experiments detailing the motivation behind each task.
 
\subsection{Models} \label{sec:model}

Each model consists of (1) an \textbf{encoder} that takes as input an image observation and outputs a latent representation and (2) a \textbf{pooling layer} that combines the latent representation into a single vector suitable to be used for the value function and policy network of an RL algorithm. We use PPO \citep{ppo} for all our experiments. Detailed information about the architecture and hyperparameters is in Appendix \ref{appx:model}.

 \textbf{Encoders.}
To investigate OCR pre-training, we evaluated three types of representations (single-vector representation, fixed-region representation, and Object-Centric Representation (OCR)) and two training regimes for each type (end-to-end learning and pre-training). Single-vector representations encode the observation to a single vector which is used in the downstream policy.
Fixed-region representations use multiple vectors to represent the image, each corresponding to a region of the observation such as a mini-patch.
For our experiments, we use the CNN feature map \citep{rn,rdrl} for end-to-end learned fixed-region representations and the mini-patch representations from pre-trained Masked AutoEncoder (MAE) \citep{mae} for pre-trained fixed-region representations.
Since the fixed-region representation can support an explicit interaction architecture, such as a Transformer encoder, it can be a stronger baseline than single-vector representations.
For end-to-end trained OCR, we use an encoder consisting of multiple CNN encoders \citep{cswm, vin}.
The pre-trained OCR methods we use in our study are described in Section \ref{sec:related_work}.
The evaluated encoders are summarized in Table \ref{tab:rep_regime_summ}.

\textbf{Pooling Layers.}
In order to use the three types of representations for RL, we implement different pooling layers for each type. For single-vector representations, an MLP (or a CNN-MLP \citep{visuomotor} in the case of SLATE-CNN) is used to obtain a single-vector representation.
For region and object-centric representations, a Transformer encoder \citep{transformer} is used.
For fixed-region representations, we add a positional embedding to identify the location of each region.
This is not needed for object-centric representations because the representations are order-invariant.
The overall architectures are shown in Figure \ref{fig:overall_archi}.

\subsection{Benchmark and Tasks}

To assess our hypotheses, particularly those inspired by the binding problem \citep{binding}, object interactions \citep{in, vin}, and relational reasoning \citep{rn, rdrl}, we designed a series of tasks and a dataset for pre-training using objects from Spriteworld \citep{spriteworld} (Figures \ref{fig:task_illustration}b-e). Our goal was to create an environment that, despite its visual simplicity, enabled OCR pre-training models to effectively segment objects in the scene into separate slots. This would minimize the impact of poor OCR quality on downstream RL performance.

To further evaluate the effectiveness of OCR pre-training in visually complex environments, we also implemented a robotic reaching task using the CausalWorld framework \citep{causalworld} (Figure \ref{fig:task_illustration}f). Comprehensive details regarding the implementation of these benchmarks, as well as comparisons with tasks utilized in prior works \citep{smorl, cobra, rdrl, visuomotor, garnelo2016towards,ke2021systematic}, can be found in Appendix \ref{appx:benchmark}.

\textbf{Dataset: } For pre-training on the 2D tasks, we generate a dataset with a varying number of objects of different shapes randomly placed in the scene.(Figures \ref{fig:task_illustration}a).
Note that this data is diverse enough to cover all four 2D tasks, so we use the same dataset for pre-training on all 2D tasks.
For 3D task from CausalWorld framework, we generate a dataset through a random policy on the task.

\textbf{Object Goal Task: } The agent (red circle), target object (blue square), and other distractor objects are randomly placed in this task.
The goal of the task is for the agent to move to the target object without touching any distractor objects.
Once the agent reaches the target object, a positive reward is given, and the episode ends.
If a distractor object is reached, the episode ends without any reward.
The discrete action space consists of the four cardinal directions to move the agent.
To solve this task, the agent must be able to extract information about the location of the target object and the objects between the agent and the target (related to the binding problem \citep{binding}).
Therefore, through this task, we can verify that the agent can extract per-object information from the representation.

\textbf{Object Interaction Task: } This task is similar to the object goal task but requires the agent to push the target to a specific location.
In Figure \ref{fig:task_illustration}c, the bottom left blue square area is the goal area.
Since the agent cannot push two objects at once, the agent must plan how to move the target to the goal area while avoiding the other objects.
Therefore, through this task, we can verify how well the agent can extract per-object information and how well the agent can reason about how the objects interact (related to the interaction between objects \citep{in,vin}).
The action space is the same as above, and the reward is only given when the agent pushes the target to the goal area.

\begin{table*}[t]
    \centering
    \begin{tabular}{@{}lccc@{}}
    \toprule
                                                      & \multicolumn{2}{c}{Training Regimes}                                               \\ \cmidrule(l){2-3}
    Representation Types                               & End-to-end Learning                             & Pre-training                 \\ \midrule
    \multirow{3}{14em}{Single-Vector Representation}    & \multirow{2}{15em}{CNN \citep{dqn}}       & VAE \citep{vae_ood}          \\
                                                      &                                                 & MAE-CLS \citep{mvp}         \\
                                                      &                                                 & SLATE-CNN \citep{visuomotor}\\\midrule
    \multirow{1}{14em}{Fixed-Region Representation}   & \multirow{1}{15em}{CNNFeat \citep{rdrl}}     & MAE-Patch                   \\\midrule
    \multirow{4}{14em}{Object-Centric Representation} & \multirow{4}{15em}{MultiCNNs \citep{cswm}}     & SLATE \citep{slate}          \\
                                                      &                                                 & Slot-Attention \citep{sa}    \\
                                                      &                                                & Slot-Attention-Large          \\
                                                      &                                                & IODINE \citep{iodine}         \\ \bottomrule
                                 
    \end{tabular}
     \caption{Summary of Evaluated Encoders}
     \label{tab:rep_regime_summ}
 \end{table*}
 
\begin{figure*}[t]
    \hspace{-8mm}
    \begin{center}
        \includegraphics[width=0.95\textwidth]{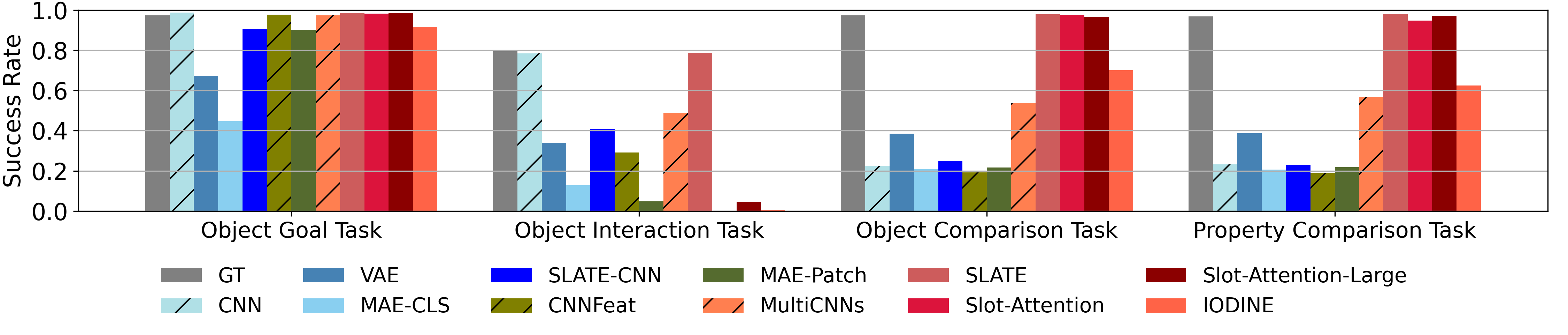}
        \caption{Success Rates for Object Goal, Object Interaction, Object Comparison, and Property Comparison Tasks. The specific representation types and training regimes used for each model are outlined in Table \ref{tab:rep_regime_summ}.
        }
        \label{fig:overall}
    \end{center}
    \vspace{-2mm}
\end{figure*}

\textbf{Object Comparison Task: } This task is designed to test relational reasoning ability. It is motivated by the odd-one-out task in cognitive science \cite{crutch2009different, stephens2008one, beatty2015prospects}, which has been previously investigated with language-augmented agents \cite{lampinen2022tell}.
To solve this task, the agent must determine which object is different from the other objects and move to it.
That is, it must find an object with no duplicates in the scene.
Unlike the object goal or object interaction tasks, the characteristics of the target object can change from episode to episode.
For example, in Figure \ref{fig:task_illustration}d, the green box is the target object in the top sample, while the blue triangle is the target object in the bottom sample.
Therefore, to know which object is the target, the agent must compare every object with every other object, which requires object-wise reasoning (related to relational reasoning \citep{rn,rdrl}).
The action space and reward structure are the same as the Object Goal Task.

\textbf{Property Comparison Task: } 
This task is similar to the Object Comparison Task, but the agent must now find the object with a \textit{property} (i.e., color or shape) that is different from the other objects.
For example, in the top sample of Figure \ref{fig:task_illustration}e, the green triangle is the target because it is the only green in the scene.
The blue triangle is the target in the bottom sample because it is the only triangle object.
Therefore, this task requires property-level comparison, not just object-level comparison (related to relational reasoning \citep{rn,rdrl}).
While OCRs are designed to be disentangled at the object level, it is not apparent how easily specific properties can be extracted and used for reasoning.
Through this task, we can verify how well OCRs can facilitate property-level reasoning.
The action space and reward structure are the same as the Object Comparison Task.

\textbf{Object Reaching Task:} Lastly, in order to evaluate the models in a more visually realistic environment, we also created a version of the Object Goal Task using the CausalWorld framework \citep{causalworld} (Figure \ref{fig:task_illustration}f).
In this environment, a fixed target object and a set of distractor objects are randomly placed in the scene.
The agent controls a tri-finger robot and must reach the target object with one of its fingers (the other two are always fixed) to obtain a positive reward and solve the task.
The episode ends without reward if the finger first touches one of the distractor objects.
The action space in this environment consists of the three continuous joint positions of the moveable finger.
We do not provide proprioceptive information to the agent, so it must learn how to control the finger from images.

\section{Experiments}

We present our experimental results and analysis as a series of questions and answers, each probing a different aspect of OCR pre-training for RL.
For each result below, we average the performance from three random seeds and every agent is trained to 2 million steps.

\begin{figure*}[t]
    \hspace{-8mm}
    \centering
    \includegraphics[width=0.95\linewidth]{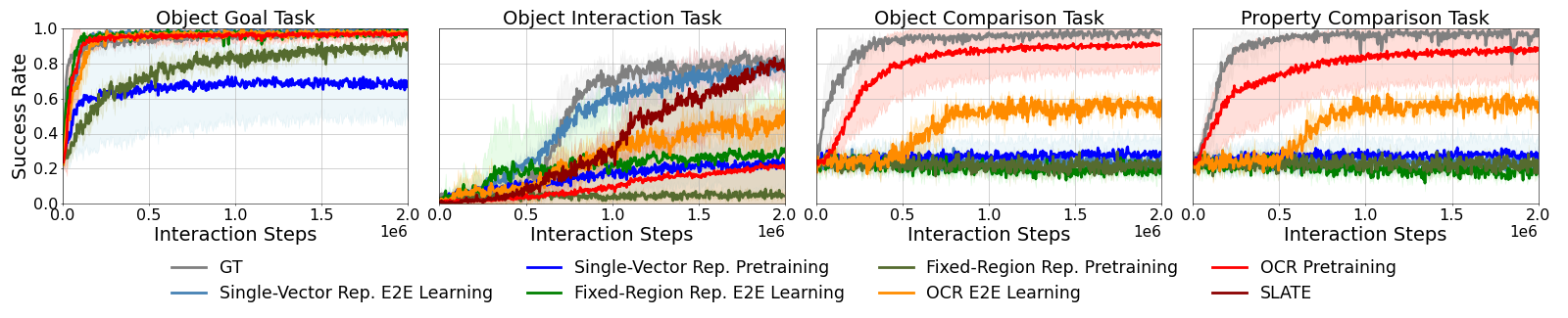}
    \vspace{-2mm}
    \caption{The comparison of success rate against the number of interaction steps with the environments. Note that SLATE is compared with baselines for the Object Interaction task, where averaged performance of OCR pre-training is hard to be compared because other OCR methods failed to solve.}
    \label{fig:overall_globalstep}
    \vspace{-3mm}
\end{figure*}

\textbf{Question 1:} \textit{Does OCR pre-training improve performance in object-centric tasks? Which types of tasks benefit the most from OCR pre-training?}
OCR pre-training represents objects through separate slots and is assumed to be beneficial for object-centric reinforcement learning (RL) tasks \citep{iodine}. Previous work has evaluated this for specific object-centric RL tasks such as goal-conditioned skill learning~\citep{smorl}. However, it remains elusive whether this representation improves performance for general object-centric tasks, which types of tasks benefit the most from OCR pre-training, and how it compares to other representations.

In Figure \ref{fig:overall_regime} and \ref{fig:overall}, we show the final Success Rates of the different models described in Table \ref{tab:rep_regime_summ} for the four synthetic tasks in our benchmark.
For the Object Goal task, where the agent must find a pre-defined goal object while avoiding distractor objects, all models achieved success rates of over 80\%, with the exception of the pre-trained single-vector representation models.
This task requires agents to extract information about the target object and the distractor objects, but it does not necessarily require modeling of interactions between objects.
This result suggests that explicit modeling of objects through OCR is not the only option for this type of task---the models with end-to-end trained single-vector representations and fixed-region representations can also be reasonable choices.
The pre-trained single-vector representation, on the other hand, seems to not be able to extract the per-object information necessary to solve the task.
We investigate this more in the response in Appendix \ref{appx:sec:more_fewer_objs}.

In the Object Interaction task, which necessitates the agent to learn object-level interactions, it is observed that only pre-trained SLATE and end-to-end learned CNN models display performance comparable to a model utilizing ground truth state, attaining a Success Rate of approximately 80\%. It is important to mention that even when using ground truth states, the agent cannot achieve perfect results, as some randomly initialized object states may be unsolvable. For instance, distractors might obstruct all possible paths to the goal position. However, it is worth noting that both the OCR model and VAE can effectively solve the task when no distractors are present, as shown in Table \ref{tab:fewer_num_objs}. It is noteworthy that other OCR pre-training models, such as Slot-Attention, Slot-Attention-Large, and IODINE, fail to accomplish this task. The divergence in performance may be due to the increased difficulty of the Object Interaction task, as it offers sparser rewards compared to others, and the fact that SLATE representations are trained with a transformer decoder, while other OCR models employ mixture-based decoders. This implies that the application of a transformer decoder in SLATE enhances the compatibility of these representations with transformer-based agent training, as opposed to other OCR models. We further explore this aspect in Appendix \ref{appx:trans_decoder_investigation}.


For the Object Comparison and Property Comparison tasks, OCR pre-training demonstrated its strengths by performing similarly to GT, with all models except IODINE. Although IODINE showed worse performance than the other OCR pre-training and GT models, it still performs better than the other baselines. End-to-end learned OCR also performed better than other baselines but was not able to fully solve the tasks (success rates were around 50\%). We hypothesize that the sparse reward structure of the task may be a contributing factor to this. Sparse rewards can hinder the E2E OCR's ability to learn good representations since they provide limited learning signals for the model to update its understanding of the problem, leading to suboptimal performance in certain cases. We also note that in contrast to E2E learned OCR, unsupervised OCR pre-training provides a stronger learning signal for discovering and representing objects, regardless of the reward sparsity.
Interestingly, fixed-region representations, namely CNNFeat and MAE-Patch, failed for the comparison tasks while also utilizing a transformer pooling layer. This is because the task requires object-level reasoning, which the fixed-region representations do not naturally provide.
These results align with the hypothesis discussed in previous works \citep{binding,van2019perspective,building} that decompositional representations are beneficial for reasoning tasks while suggesting that the level of decompositionality of representations is also important.
Another interesting point is that OCRs were not necessarily disentangled at the property level but are effectively utilized to solve property-level comparisons. We hypothesize that the transformer pooling layer plays a critical role in correctly extracting property-level information. We investigate this hypothesis more in Appendix \ref{appx:trans_pooling_property_comparison}.

In conclusion, OCR pre-training does not always provide better performance for every object-centric task, but for relational reasoning tasks, it demonstrates better performance when compared to other diversely trained representations.

\textbf{Question 2:} \textit{Does OCR pre-training improve sample efficiency in object-centric tasks?}
Analyzing the sample efficiency of OCR pre-training compared with other methods will give us insights into whether or not the learned representations are appropriate for the downstream tasks.
As shown in Figure \ref{fig:overall_globalstep}, OCR pre-training generally demonstrates improved sample efficiency compared to other pre-training methods.
The exception is for the non-SLATE OCR pre-training methods for the Object Interaction task that failed to solve the task.
This shows that when compared to pre-training with single-vector or fixed-region representations, the pre-trained object-centric representations are more suitable for these tasks.
When compared with the end-to-end trained methods, we see that OCR pre-training also has better sample efficiency for the comparison tasks but worse sample efficiency for the object interaction task.
These results suggest that, as with performance improvement, OCR pre-training may not always improve sample efficiency for every object-centric task, but it does enhance sample efficiency for tasks where the relationship between objects is important.

\begin{figure*}[t]
    \hspace{-10mm}
    \begin{center}
    \includegraphics[width=0.95\linewidth]{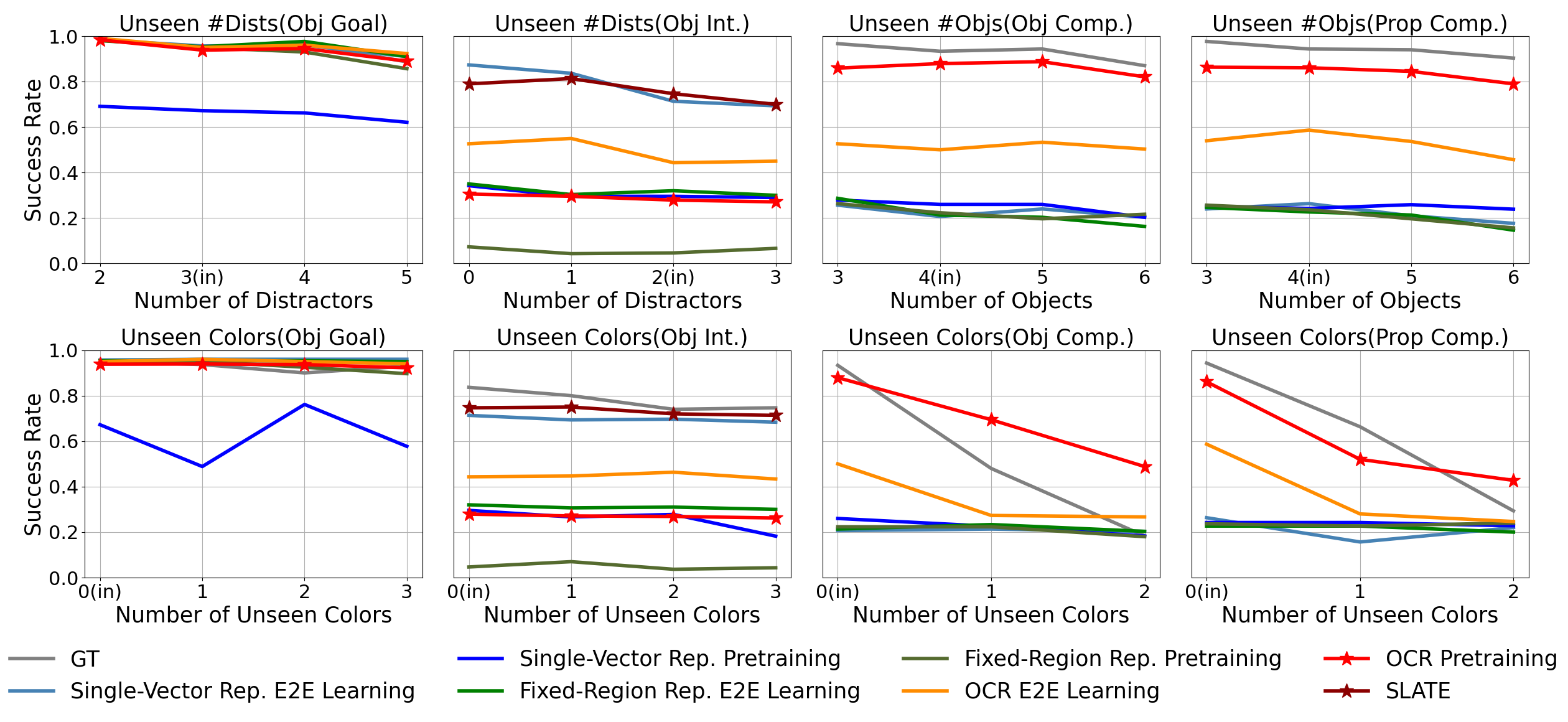}
    \end{center}
    \vspace{-2mm}
    \caption{Generalization performance for the out-of-distribution settings. The in-distribution setting is denoted by ``(in)''. The \textbf{top row} shows the success rate for unseen numbers of objects, and the \textbf{bottom row} shows the success rate for unseen object colors. SLATE is compared for the Object Interaction task. OCR pre-training and SLATE are highlighted through markers. Note that GT is not evaluated for the unseen number of objects test on the Object Interaction task as the MLP pooling is used for the task and cannot be applied to unseen objects.}
    \label{fig:ood}
    \vspace{-3mm}
\end{figure*}

\textbf{Question 3:} \textit{Does OCR pre-training help in generalization of agents?} \label{sec:ood_exp}
OCR has been shown to have strong generalization capabilities due to its object-wise modular representations, particularly when dealing with out-of-distribution data such as unseen numbers or combinations of objects \citep{generalization_ocr,sa,iodine,slate}.
Agents that utilize explicit interaction networks, such as Transformers \citep{rdrl} or Linear Relational Networks \citep{lrn}, have also demonstrated good generalization performance in policy learning.
It can be hypothesized that OCR pre-training in conjunction with explicit interaction networks could further improve generalization robustness, but this has not yet been thoroughly investigated. In the following, we will investigate the generalization capabilities of OCR pre-training in the context of two different types of distribution shifts: unseen number of objects and unseen types of objects.

First, we investigated the effect on agent performance when the number of objects differs from that on which the agent was trained.
The results are shown in the top row of Figure \ref{fig:ood}.
For all tasks, OCR pre-training generally maintained good performance, although success rates on Object Interaction tasks were low.
For the Object Interaction task, SLATE shows comparable generalization performance.
However, it is worth noting that other methods, such as GT or CNN, also demonstrated comparable generalization capabilities.
This is not surprising for the Object Goal and Interaction tasks, since the model must extract the target object from the observations.
For comparison tasks, however, increasing the number of objects can create unseen patterns such as the ones in Figure \ref{fig:task_illustration}e, and the agent must compare each object to all other objects to find the odd one.
The transformer pooling layer can handle this pairwise comparison, which is why OCR pre-training and GT perform well when scaling to more objects.

Next, we evaluate the agent's performance when presented with object colors not seen during training. The experimental details, such as which colors were changed, are described in Appendix \ref{appx:ood_unseen_color_detail}. The results are shown in the bottom row of Figure \ref{fig:ood}.
For the Object Goal and Interaction tasks, we change the color of the distractors. The target object remains the same, so we can infer that this distribution shift does not affect performance if the agent can correctly extract the target object. As expected, model performance remains relatively stable regardless of the number of unseen distractors.
For the comparison tasks, however, the agent must compare each object, and unseen colors can negatively impact performance.
As expected, the performance of every model that performs better than random chance significantly decreased. Especially, the GT success rate drops from almost 100\% to 2-30\%. This is likely due to the unseen index (object type is represented as an integer index in the GT state) being critical to infer the correct action. On the other hand, except for the Property Comparison task with one unseen color, OCR pre-training demonstrates more robust performance than GT.

\textbf{Question 4:} \textit{Does OCR pre-training work well in visually complex environments where segmentation is difficult?} 
In order to evaluate the effectiveness of OCR pre-training in visually complex environments where segmentation is challenging, we conducted experiments on the Object Reaching task using the SLATE model and several baselines (GT, CNN, and VAE). The results of the segmentation performed by SLATE, as shown in Figure \ref{fig:cw_seg}, demonstrate that it is not perfect and sometimes splits multiple objects between slots and does not accurately capture the robotic finger.

\begin{figure}[h]
  \begin{center}
   \includegraphics[width=0.7\linewidth]{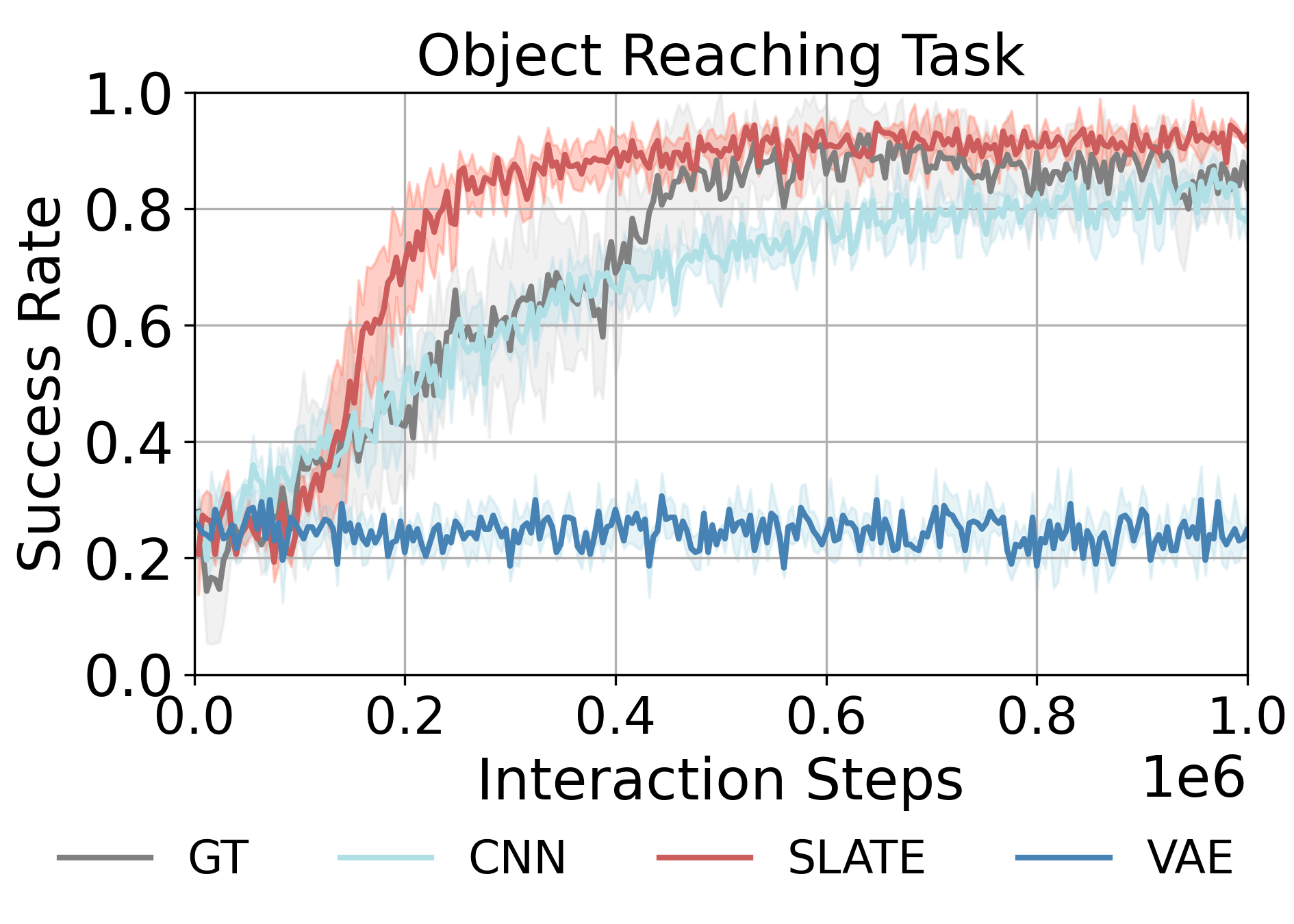}
  \end{center}
    \vspace{-5mm}
  \caption{Success rates for the Object Reaching Task.}
  \label{fig:cw}
    \vspace{-1mm}
\end{figure}

However, as illustrated in Figure \ref{fig:cw}, the agent utilizing SLATE demonstrated superior sample efficiency and converged success rate compared to the other methods. Although this task does not explicitly require reasoning among the objects, it is still crucial for the agent to learn to avoid touching the distractor objects before the target object. This result suggests that the conclusions from the experiments in visually simple environments can potentially be generalized to more complex environments.

\textbf{Question 5:} \textit{Which OCR model is better for RL?} 
From the results presented in Figures \ref{fig:overall} and \ref{appx:fig:each_model_walltime2}, it is clear that the SLATE model performs the best in terms of overall performance on the tasks evaluated in this study. In contrast, the IODINE model demonstrated slower computational times and inferior performance across all tasks. Slot-Attention performed similarly to the SLATE model on Object Goal, Object Comparison, and Property Comparison tasks, but failed to effectively solve the Object Interaction task, even when utilizing a more extensive architecture (Slot-Attention-Large). 

\begin{table}[h]
    \begin{center}
    \begin{tabular}{@{}lcccc@{}}
    \toprule
             & \multicolumn{4}{c}{Tasks}                               \\ \cmidrule(l){2-5}
    Models   & Goal         & Int.         & Obj. Com.    & Pro. Com.  \\ \midrule
    SLATE    &\textbf{0.985}&\textbf{0.787}&\textbf{0.979}&\textbf{0.98} \\
    SLATE-MLP&0.9           &0.03          &0.238         &0.229         \\ \bottomrule
    \end{tabular}
    \caption{Success Rate comparison between Transformer and MLP pooling layers}\label{tab:slate_mlp}
    \end{center}
\end{table}

\textbf{Question 6:} \textit{How does the choice of pooling layer affect task performance?}
In this study, the transformer pooling layer is used for the OCR pre-training models due to its permutation invariance and ability to explicitly model interactions between the slots, which are important properties for relational reasoning tasks. To evaluate the effect of the pooling layer on task performance, an ablation study is conducted, where the MLP pooling layer is applied to the SLATE (referred to as SLATE-MLP). The results, as shown in Table \ref{tab:slate_mlp}, indicate that the use of the MLP pooling layer resulted in inferior performance on all tasks, with complete failure to solve the interaction and comparison tasks. Interestingly, the SLATE-MLP is still able to achieve good performance on the Object Goal task. This may be because the task is easier than others, the target object can still be extracted from the MLP, and the interaction between objects is not very important to solve that task.

\section{Conclusion and Discussion}

In this paper, we investigated Object-Centric Representation for RL tasks. To do this, we empirically evaluated the hypotheses shown or suggested from previous works \citep{smorl,binding,van2019perspective,building,rdrl, load,lrn} on our new benchmark with diverse types of representations. We found more specific conditions to satisfy the hypotheses through this empirical investigation. OCR pre-training does not always provide sample efficiency but is efficient for relational reasoning tasks. OCR pre-training is not always better than other methods for out-of-distribution tasks also, but it shows better generalization performance for unseen objects when compared with GT.

In addition, SLATE shows comparable performance to GT for visually complex tasks where OCR cannot segment objects correctly. We also studied ablations for the pooling layer. Other interesting questions were also considered, but due to space limitation, they are discussed in the Appendix \ref{appx:additional_res}.

Although our benchmark covers several important object-centric tasks, such as object interaction and relational reasoning, it can seem very synthetic. We chose those visually simple scenes to ensure the downstream RL performance is not affected by poor segmentation quality. It allows us to probe more specific aspects of the reinforcement learning task to assess where OCR pre-training is most beneficial. In order to investigate the case where segmentation quality is not perfect, we also ran experiments on the robotics Object Reaching Task, which we discuss in \textbf{Question 4}. Investigating OCR in more complex and realistic environments is a promising direction for future work, especially as unsupervised OCR models continue to improve. Furthermore, in addition to scene complexity, there are other aspects of agent learning that can benefit from OCR, such as partially observable environments or tasks that require exploration.
These are good candidates to extend the benchmark in the future.

Lastly, we hope our benchmark can help evaluate OCR models in the context of agent learning, in addition to the previously standard metrics such as segmentation quality and property prediction accuracy. Further discussion is in Appendix \ref{appx:discussion}.

\section*{Acknowledgement}
This work is supported by Brain Pool Plus (BP+) Program (No. 2021H1D3A2A03103645) and Young Researcher Program (No. 2022R1C1C1009443) through the National Research Foundation of Korea (NRF) funded by the Ministry of Science and ICT.
The work is also supported by Electronics and Telecommunications Research Institute (ETRI) grant funded by the Korean government. [23ZR1100, A Study of Hyper-Connected Thinking Internet Technology by autonomous connecting, controlling, and evolving ways] We thank to Andrea Dittadi for valuable discussions, and JS would like to thank SAP and Dr. Han's team for their support and dedicated medical care.

\bibliography{example_paper}
\bibliographystyle{icml2023}

\newpage
\appendix
\onecolumn

\section{Model Details} \label{appx:model}
In this section, we introduce the architectural details about the models we used in our experiments.

\subsection{Encoder}

\subsubsection{Ground Truth State (GT)}

Ground Truth (GT) states are used as a baseline for comparison in the experiments. In 2D tasks, the GT state is represented as a matrix of the number of objects $\times$ 5. Each object is represented by a 5-dimensional vector, consisting of the COLOR index, SHAPE index, SIZE index, x-coordinate, and y-coordinate. The color, shape, and size indices are chosen from pre-specified sets. The pre-specified color set includes [blue, green, yellow, red, cyan, pink, brown], and the pre-specified shape set includes [square, triangle, star\_4, circle, pentagon, hexagon, octagon, star\_5, star\_6, spoke\_4, spoke\_5, spoke\_6]. The size of the objects ranges from [0.15, 0.22]. For tasks that use the MLP pooling layer, an additional two-layer MLP with 32 units and rectified linear unit (ReLU) activation is applied on top of the GT per-object state.

The ground truth state for the 3D Object Reaching task is represented by a concatenation of the robot state and object states. The robot state is composed of 37 dimensions, including joint positions, velocities, and end effector positions. Each object is represented by 9 dimensions, including its cartesian position, size, and color (RGB). These states are concatenated together and an additional dimension is added to indicate whether the slot corresponds to the robot arm or an object. The final representation consists of 5 slots, each with 37 dimensions.

\subsubsection{CNN, CNNFeat and MultiCNNs}

CNN, CNNFeat and MultiCNNs are based on the same CNN architecture for encoding the observations. The architecture is similar to the one used in \citep{mnih2015human}, and the implementation is sourced from the Stable Baselines3 library \citep{sb3}. CNNFeat is obtained by removing the MLP layers after CNN encoding. MultiCNNs is implemented by using multiple CNN models with non-shared parameters.

\subsubsection{VAE}

The VAE model used in this study employs a multi-block CNN architecture for both the encoder and decoder. The encoder block comprises of four CNN layers, with channel sizes of [64, 64, 64, 64], kernel sizes of [2, 1, 1, 1], and strides of [2, 1, 1, 1]. Padding is set to zero for all layers, and ReLU activation is used. The decoder block also consists of four CNN layers, with channel sizes of [64, 64, 64, 64*4], kernel sizes of [3, 1, 1, 1], strides of [2, 1, 1, 1], and paddings of [1, 0, 0, 1]. ReLU activation is used for all layers.

The encoder is composed of four encoder CNN blocks and one CNN layer, with channel size of 64, kernel size of 1, stride of 1, padding of 0, and no activation function is applied. The CNN feature map is flattened and passed through linear layers to obtain the mean and variance of the latent variable, with a size of 256.

To decode, the latent variable is passed through a linear layer to match the size of the CNN feature map. The decoder includes four decoder blocks with a pixel shuffle function between blocks, and one CNN layer, with channel size equal to the observation channel size, kernel size of 1, stride of 1, padding of 0, and no activation function is applied, to reconstruct the observations.

Additional hyperparameters include a learning rate of $0.0001$, a weight for the KL-term of 5, and a batch size of $128$.

\subsubsection{MAE}

The Multi-modal Auto-Encoder (MAE) encoder architecture is based on the Vision Transformer Base (ViT-Base) architecture \citep{vit}. The masking ratio used during pre-training was set to 0.5, which is lower than the best configuration reported in \citep{mae}. This decision was made as the observations in our task consist of multiple small objects, and a high masking ratio can result in the masking of entire objects. The patch size used during pre-training was set to 16, resulting in the same number of patches as the CNNFeat model. The pre-training of the MAE encoder was conducted using a batch size of 128, a learning rate of 0.001, and a weight decay of 0.05.

\subsubsection{IODINE}

The IODINE model is based on the architecture reported in the original paper \citep{iodine} for the CLEVR dataset. The only modification made to the original configuration is the use of $\sigma=0.35$ when reconstructing the image.

\subsubsection{SLATE-CNN/Slot-Attention/Slot-Attention-Large/SLATE}

The SLATE model uses a CNN encoder, similar to the architecture used in Slot-Attention, as reported in \citep{slate}. The Slot-Attention model follows the architecture described in \citep{sa}, and the Slot-Attention-Large model utilizes a larger architecture with the same size as the SLATE model. The number of slots used in the model varies depending on the task. Detailed hyperparameter settings can be found in Tables \ref{appx:tab:slate} and \ref{appx:tab:sa}.

\subsection{Pooling and Policy}

In this study, we utilized two types of pooling layers: the Transformer encoder \citep{transformer} and Multi-Layer Perceptron (MLP). The transformer pooling layer used a hidden size of 128 and 8 heads. The number of layers of the transformer pooling layer varied across tasks and encoders, with the optimal number chosen among 1 or 3. The MLP pooling layer consisted of two linear layers with a size of 128 and ReLU activation function.

For the policy algorithm, we used Proximal Policy Optimization (PPO) \citep{ppo} with a learning rate of 0.0003. Additional configurations were tuned across tasks and models. The steps per training were selected from 2048 or 8192, and the coefficient for the entropy term was selected from 0, 0.01, 0.03, 0.05, or 0.1. The policy was trained using the Stable Baselines3 library \citep{sb3}, with trajectories collected through 4 environments.

\section{Benchmark} \label{appx:benchmark}

Our benchmark comprises of 2D tasks from the Spriteworld \citep{spriteworld} and a 3D task from the CausalWorld \citep{causalworld}.

\subsection{2D Tasks} \label{appx:benchmark_2d}

In the 2D tasks, the observation size and channels are 64 and 3, respectively. The object size is represented as a percentage of the observation size. There is no occlusion between objects and the agent is always represented as a red ball. At the beginning of each episode, the agent's position is fixed at the center of the observation. The action set consists of four actions: move up, move down, move left, and move right. The agent moves 0.05 units in the chosen direction at each action. The objects are randomly distributed in the scene and their characteristics are sampled from a pre-specified set in accordance with the rules of each task. For the tasks, the size of every object is 0.15.

For pre-training, a non-task-specific dataset is used. The dataset is comprised of scenes with randomly distributed objects. The number of objects in each scene is 5 and the object color is one of [blue, green, yellow, red] and shape is one of [square, triangle, star\_4, circle]. The size is one of [0.15, 0.22]. The minimum distance between objects is 0.15, thus occlusion can happen when the object size is 0.22. The number of scenes used for training and validation are 1 million and 100,000, respectively.

In the Object Goal Task, the sets of shapes and colors that are used are [square, triangle, star\_4], and [blue, green, yellow, red], respectively. The target object is always a blue square, and only one target object is present in the environment at a time. The other objects are randomly generated.

In the Object Interaction Task, the color set used is [blue, green, yellow, red] and the shape is fixed as square. The target object is always a blue square. To ensure the task is solvable, all objects are positioned a distance from the walls that is at least the size of the object, allowing the agent to push the target in any direction.

In the Object Comparison Task, the shape and color sets used are [square, triangle] and [blue, green], respectively. There must be a single unique object in the environment, with the other objects being randomly generated according to this rule.

In the Property Comparison Task, the shape and color sets used are the same as in the Object Comparison Task. The task requires that there is only one unique property present in the environment, such as only one square or only one blue object, with the other objects being randomly generated according to this rule.

\subsection{3D Object Reaching Task}

In the 3D Object Reaching task, the task environment consists of a tri-finger robotic arm and multiple objects with different colors and sizes. The agent's task is to reach and manipulate the target object, which is always blue, among the randomly generated objects. The observations are rendered images with a resolution of 64x64 and 3 channels. The objects are cubes with a color set of [blue, green, yellow, red].
The actions are limited to the manipulation of the third finger of the robotic arm, while the other two fingers are fixed in the upright position. The agent starts each episode with all fingers in the upright position.
For pre-training, a dataset of 1 million observations is collected through a random policy is utilized.

\subsection{Comparison of Tasks in Our Benchmark with Previous Works}

In this section, we discuss how tasks in our benchmark relate to tasks from previous works in order to identify any tasks we might have overlooked. In \citep{smorl}, goal-conditioned tasks were used to validate their approach, wherein the agent had to push objects according to a specified goal. This task is related to our object interaction task, as it investigates object interactions. In \citep{cobra}, three tasks—Goal-Finding, Sorting, and Clustering—were employed to validate their model. The Goal-Finding task required the agent to bring a set of target objects, identified by a feature such as color, to a hidden location while ignoring distractor objects. This task can be associated with our object goal task since it involves extracting target object knowledge from a scene containing multiple objects. The Sorting task, which requires the agent to move objects to a goal location based on their color, is also related to our object goal task. The Clustering task demands that the agent groups objects by their color, necessitating the comparison of each object's property and subsequent grouping. Consequently, it can be linked to our property comparison task.

It is worth noting that in \citep{cobra}, only color was used for property-level comparisons, while our study also tested shape. Color is a lower-level visual feature, whereas shape is a higher-level category description (e.g., finding color could be accomplished by searching pixel knowledge without identifying objects, but comparing shapes requires the model to identify the entire object). In earlier studies where OCR pre-training was not utilized but decomposed representations were employed for RL, the Box-World task in \citep{rdrl} can be connected to our object comparison task, as it requires linking the same object. The robotic task used in \citep{visuomotor} demanded learning about object interactions, which is related to our object interaction task. The symbolic task in \citep{garnelo2016towards} can be associated with our object goal task. In \citep{ke2021systematic}, two environments, physics and chemistry environments, are evaluated. In the physics environment, weighted-block pushing is assessed. Regarding the evaluation of object interactions, it could be related to the object interaction task while interactions between weighted blocks are evaluated. Since the weights were represented through color, it can be seen as an evaluation of interaction and binding problems. In the chemistry environment, they attempted to evaluate the discovery of various causal graphs, which is not evaluated in our benchmark. It is important to note that we did not link complex tasks such as the Starcraft benchmark in \citep{rdrl}, as they might require multiple aspects we previously discussed, such as reasoning about object interactions and relational reasoning, simultaneously.

In summary, while our benchmark does not include tasks that require discovering complex causal graphs, it covers a wide range of tasks from previous works \citep{smorl,cobra,rdrl,visuomotor,garnelo2016towards,ke2021systematic}, demonstrating that our benchmark encompasses a variety of challenges.

\section{Experiment Details}

\subsection{The unseen color evaluation} \label{appx:ood_unseen_color_detail}

In order to evaluate the generalization capabilities of the agent to unseen objects, we use unseen colors in our experiments. Specifically, for the Object Goal and Interaction tasks, we alter the colors of the distractor objects while keeping the target object (blue square) constant. The in-distribution color set used in these tasks is [blue, green, yellow, red], and it is progressively modified as follows: [blue, green, yellow, pink] $\rightarrow$ [blue, green, brown, pink] $\rightarrow$ [blue, cyan, brown, pink].

Similarly, for the Comparison tasks, we change the color of any object. The in-distribution color set used in these tasks is [blue, green], and it is progressively modified as follows: [blue, pink] $\rightarrow$ [cyan, pink]. This allows us to evaluate the agent's ability to generalize to new and unseen combinations of object properties.

\section{Additional Results} \label{appx:additional_res}

\begin{figure}[h]
    \hspace{-8mm}
    \centering
    \includegraphics[width=0.97\linewidth]{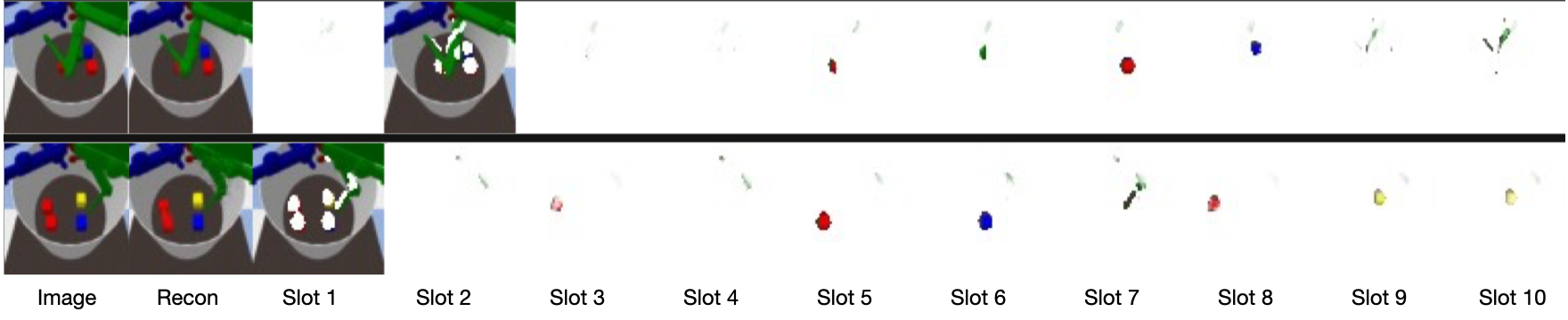}
    \caption{SLATE segmentation on the Object Reaching Task}
    \label{fig:cw_seg}
    \vspace{-3mm}
\end{figure}

\begin{table}[h]
\centering
\begin{tabular}{ccccccc}
\toprule
                    & \multicolumn{3}{c}{Unseen Shapes(Obj Comp.)} & \multicolumn{3}{c}{Unseen Shapes(Prop Comp.)} \\\cmidrule(l){2-4} \cmidrule(l){5-7}
Models              & 0(in)         & 1             & 2            & 0(in)         & 1             & 2             \\ \midrule
GT                  & 0.933         & 0.373         & 0.213        & 0.943         & 0.480         & 0.153         \\
CNN                 & 0.207         & 0.213         & 0.247        & 0.263         & 0.180         & 0.207         \\
CNNFeat             & 0.213         & 0.237         & 0.240        & 0.227         & 0.210         & 0.190         \\
MultiCNNs           & 0.500         & 0.490         & 0.530        & 0.587         & 0.603         & 0.610         \\
VAE                 & 0.383         & 0.353         & 0.373        & 0.400         & 0.393         & 0.393         \\
MAE-CLS             & 0.197         & 0.250         & 0.220        & 0.140         & 0.187         & 0.193         \\
SLATE-CNN           & 0.200         & 0.257         & 0.217        & 0.187         & 0.200         & 0.217         \\
MAE-Patch           & 0.223         & 0.197         & 0.257        & 0.237         & 0.230         & 0.217         \\
SLATE               & \textbf{0.950}& 0.557         & 0.447        & 0.927         & 0.603         & 0.527         \\
SlotAttention       & 0.937         & 0.740         &\textbf{0.573}& 0.920         & 0.620         & \textbf{0.627}\\
SlotAttention-Large & 0.930         & \textbf{0.820}&\textbf{0.573}& \textbf{0.957}& \textbf{0.720}& 0.620         \\
IODINE              & 0.700         & 0.503         & 0.503        & 0.640         & 0.553         & 0.533         \\ \bottomrule
\end{tabular}
    \caption{The performances for unseen shapes on Object and Property Comparison tasks.}
    \label{appx:tab:unseen_shape}
\end{table}

\subsection{How does OCR pre-training generalization performance for unseen shapes on the Object and Property Comparison tasks?}

The generalization performance of the agents when the distribution shift occurs in the shape property of the objects is evaluated in this study. Specifically, we focus on the Object and Property Comparison tasks. The in-distribution shape set used in these tasks is [square, triangle], and it is progressively modified as follows: [star\_5, triangle] $\rightarrow$ [star\_5, spoke\_4]. The results, presented in Table \ref{appx:tab:unseen_shape}, indicate that models pre-trained with OCR show more robust performance for unseen shapes compared to models pre-trained with GT. Interestingly, MultiCNNs show good generalization performance for unseen shapes, while their performance is worse for unseen color generalization tests, as shown in Table \ref{appx:tab:unseen_color2}. We posit that this may be due to the fact that MultiCNNs are solving the tasks using color information alone, hence, its performance for the in-distribution case is around 50\% and is not affected by shape distribution shifts.

\begin{table}[h]
\centering
    \begin{tabular}{@{}lccc@{}}
    \toprule
         &              & \multicolumn{2}{c}{Success Rate}   \\ \cmidrule(l){3-4}
Setting  &  Model     & ID for Pretraining, OOD for Agent & OOD for Pretraining and Agent \\ \midrule
  unseen shapes       &  SLATE &  $\mathbf{0.663} \pm 0.049$  &  $0.527 \pm 0.16$    \\ 
 \midrule      
  unseen colors       &  SLATE &  $ 0.567 \pm 0.021$  &  $\mathbf{0.590} \pm 0.04$    \\ 
                                 \bottomrule
    \end{tabular}
    \caption{The generalization performance when the setting is in-distribution for pre-training and out-of-distribution for agent learning.}\label{appx:tab:id_ocr_ood_agent}
\end{table}

\subsection{How does out-of-distribution generalization differ when the OCR pre-training environment is in-distribution but the agent environment is out-of-distribution?} \label{appx:id_ocr_ood_agent}

In this study, we evaluate the generalization performance of agents when the environment is in-distribution for the pre-training dataset used for the encoders but out-of-distribution for the agent. The in-distribution color set used in these tasks is [blue, green], and it is progressively modified as follows: [blue, yellow] $\rightarrow$ [red, yellow]. Specifically, we focus on the SLATE model for the Property Comparison task and the results, as presented in Table \ref{appx:tab:id_ocr_ood_agent}, indicate that the agent performs better in an environment that is in-distribution for the pre-training dataset but out-of-distribution for the agent, compared to an environment that is out-of-distribution for both. This is observed for the unseen shape condition, where the SLATE model shows better performance when the shape is in-distribution for the pre-training but out-of-distribution for the agent. However, for the unseen color condition, the performance on the two tests is similar. This discrepancy may be attributed to the fact that shape information is not as easily represented through lower-level cognition such as pixel-level representations, whereas color representation is more easily generalized even when it is unseen in the pre-training dataset. However, the unseen color combination may not be well solved through the agent even though the color itself was seen in the pre-training.

\begin{figure}[h]
    \hspace{-8mm}
    \begin{center}
    \includegraphics[width=0.95\linewidth]{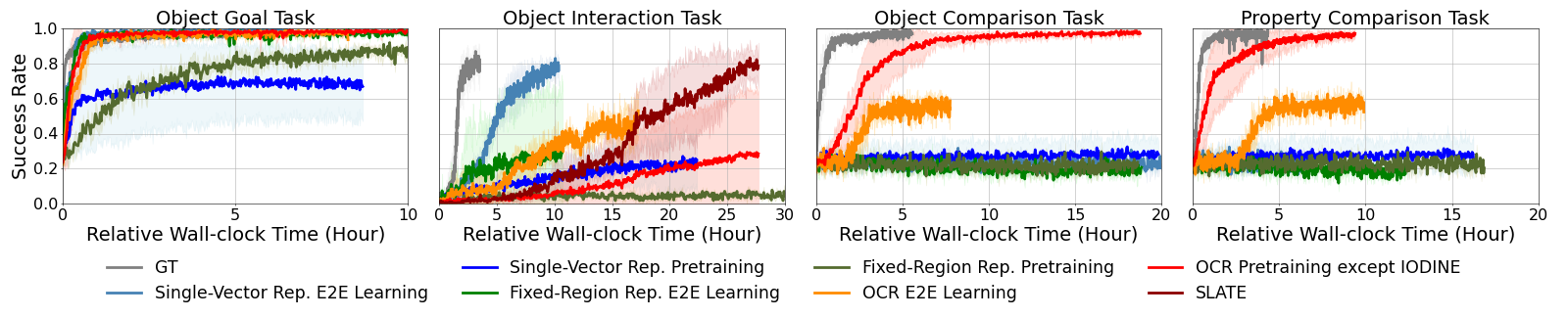}
    \vspace{-3mm}
    \end{center}
    \caption{Success rate comparison vs. wall-clock time. IODINE is not compared because IODINE is too slow to compare with other methods.}
    \label{fig:overall_wallclock}
    \vspace{-5mm}
\end{figure}

\subsection{Is OCR pre-training more efficient than end-to-end training in terms of wall-clock time?}
\begin{figure}[h]
    \centering
    \includegraphics[width=0.90\linewidth]{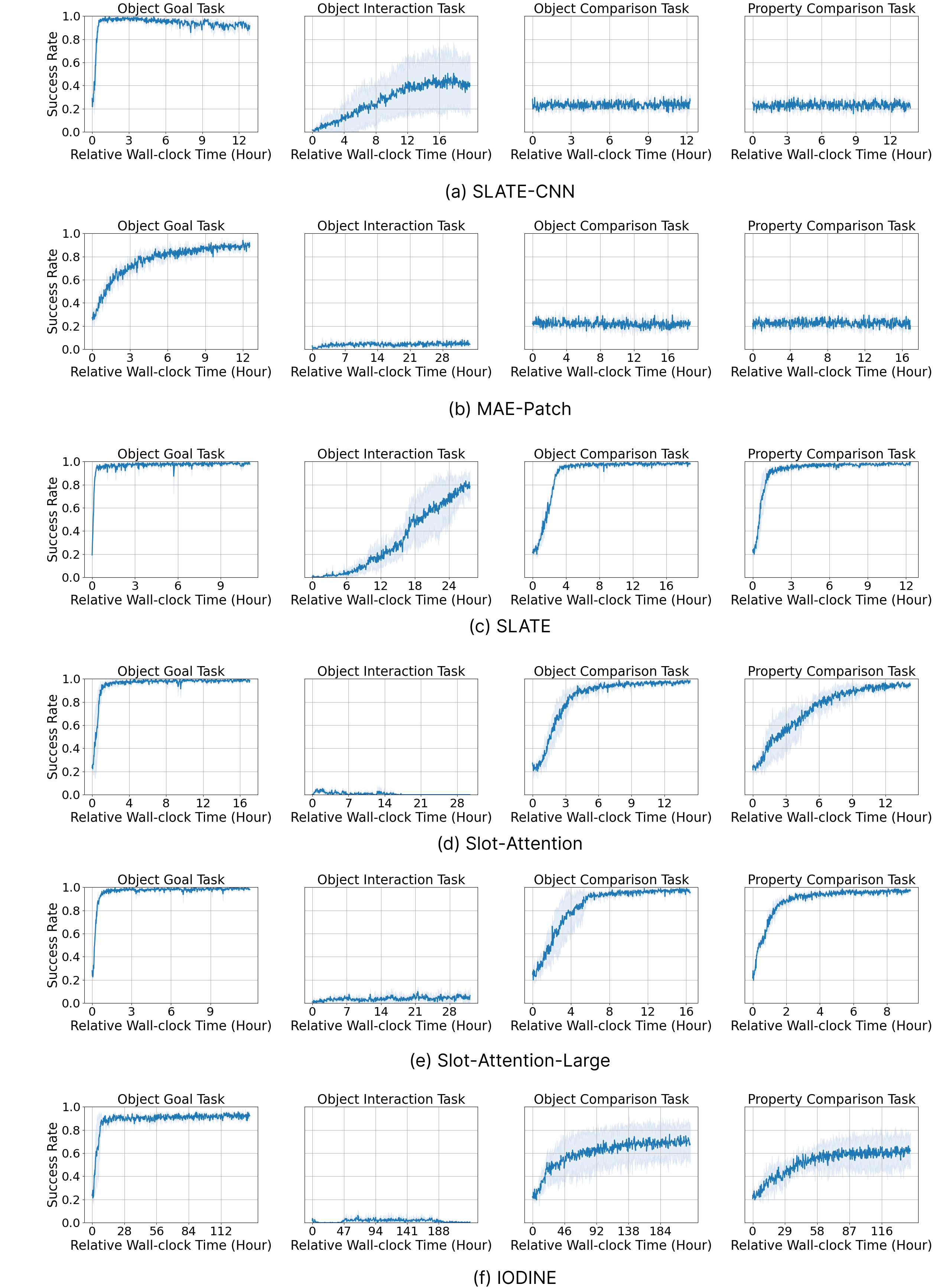}
    \caption{SLATE-CNN / MAE-Patch / SLATE / Slot-Attention / Slot-Attention-Large / IODINE performances for relative wall-clock time}
    \label{appx:fig:each_model_walltime2}
\end{figure}

The question of whether OCR pre-training is more efficient than baselines in terms of wall-clock time is of particular interest, as OCR models typically require more computation than other types of representations. In order to address this question, we compare the wall-clock time of OCR pre-training with other methods, as shown in Figure \ref{fig:overall_wallclock}.
Note that due to IODINE's excessive computation time, we plot it separately in Figure \ref{appx:fig:each_model_walltime2} to keep the chart legible.
These results align with the findings regarding sample efficiency in \textbf{Question 2}.
OCR pre-training is efficient for comparison tasks, while it is comparable to end-to-end learning methods for the Object Goal task.
However, for the Object Interaction task, the gap between end-to-end learned CNN and SLATE is much larger than that observed in \textbf{Question 2}, due to the computational demands of SLATE.

\begin{table}[h]
    \centering
    \begin{tabular}{@{}lccc@{}}
    \toprule
         &              & \multicolumn{2}{c}{Models}   \\ \cmidrule(l){3-4}
Task  &  \#Objs     & SLATE  & CNN  \\ \midrule
  \multirow{2}{6em}{Object Goal}       &  4 &  $0.979 \pm 0.01$  &  $\mathbf{0.985} \pm 0.01$    \\ 
                                 &  6   &  $\mathbf{0.95} \pm 0.00$     &  $0.746 \pm 0.00$   \\ \bottomrule
    \end{tabular}
    \caption{Comparison when more number of objects}\label{tab:more_num_objs}
\end{table}
\begin{table}[h]
    \centering
    \begin{tabular}{@{}lccc@{}}
    \toprule
         &              & \multicolumn{2}{c}{Models}   \\ \cmidrule(l){3-4}
Task  &  \#Objs     & SLATE  & VAE  \\ \midrule
  \multirow{2}{6em}{Object Goal}       &  1 &  $0.997 \pm 0.01$  &  $\mathbf{0.999} \pm 0.00$    \\ 
                                 &  3   &  $\mathbf{0.992} \pm 0.01$     &  $0.686 \pm 0.02$   \\ 
 \midrule      
  \multirow{2}{6em}{Object Int.}       &  1 &  $ \mathbf{0.99} \pm 0.01$  &  $0.971 \pm 0.01$    \\ 
                                 &  3   &  $\mathbf{0.787} \pm 0.03$     &  $0.345 \pm 0.08$   \\      
                                 \bottomrule
    \end{tabular}
    \caption{Comparison when fewer number of objects}\label{tab:fewer_num_objs}
 \end{table}
 
\subsection{Does OCR pre-training work better than the baselines in environments with more objects? What happens if there are fewer objects?} \label{appx:sec:more_fewer_objs}

The binding problem in neural networks refers to the difficulty in representing multiple objects as distinct entities when they are encoded into a single vector representation \citep{binding}. OCRs, on the other hand, provide a scalable solution to this problem, as they can represent multiple objects independently.

To investigate the effect of the number of objects on the performance of OCR pre-training, we evaluated the SLATE and CNN models on the Object Goal task while increasing the number of objects in the scene. The results, presented in Table \ref{tab:more_num_objs}, show that as the number of objects increases, the performance of both models decreases. However, the performance degradation of the CNN model is much greater than that of the SLATE model.

We also evaluated the performance of VAE and SLATE when the number of objects in the environment is reduced. The results in Table \ref{tab:fewer_num_objs} show that with fewer objects in the environment, both VAE and SLATE performed better, with the difference being more pronounced for the VAE model. Notably, when there is only one object in the environment, the VAE model showed similar performance to the SLATE model on both the Object Goal and Object Interaction tasks. These results provide evidence of the binding problem in single-vector representations and highlight the scalability and robustness of OCR pre-training in environments with varying numbers of objects.

\begin{table}[h]
\centering
\begin{tabular}{ccccc}
\toprule
                    & Transformer pooling & Deep Sets pooling & Relational Network pooling & MLP pooling    \\ \midrule
Ground Truth State  & 0.027               & 0.697             & 0.317                      & \textbf{0.795} \\
SLATE               & \textbf{0.787}      & 0.103             & 0.093                      & 0.03           \\
SlotAttention-Large & 0.047               & \textbf{0.273}    & 0.087                      & - \\ \bottomrule            
\end{tabular}
    \caption{Object Interaction Task performance for Ground Truth State, SLATE, and SlotAttention-Large with additional pooling layers, Deep Sets \citep{deepset} and Relational Network \citep{rn}.}\label{appx:tab:additional_results_interaction_task}
\end{table}

\subsection{Does the transformer decoder in SLATE truly enhance compatibility with transformer-based agent training?} \label{appx:trans_decoder_investigation}

To further examine the hypothesis that the use of a transformer decoder in SLATE improves the compatibility of its representations with transformer-based agent training, we conducted additional experiments. We implemented two different pooling layers, Deep Sets \citep{deepset} and Relational Networks \citep{rn}, for OCRs and ground truth states in targeted controlled experiments, focusing on the Object Interaction Task. We should note that we did not test the MLP pooling layer for SlotAttention-Large, as it was already demonstrated in Question 6 that the MLP pooling layer is not suitable for OCRs due to the orderless nature of slots.

Our results in Table \ref{appx:tab:additional_results_interaction_task} reveal that the Deep Sets pooling layer performs best for the SlotAttention-Large model with OCRs, while the MLP pooling layer exhibits the best performance for ground truth states. Interestingly, for SLATE, the Transformer pooling layer demonstrates the highest performance, and when using the Deep Sets pooling layer, its performance is lower than that of SlotAttention-Large. This lends further support to the notion that employing a transformer decoder in SLATE may enhance the compatibility of OCRs with transformer-based agent training.

To ensure a fair comparison, we adhered to the hyperparameters used in \citep{deepset,rn} for the Deep Sets and Relational Network pooling layers.

\begin{table}[h]
\centering
\begin{tabular}{cccc}
\toprule
                    & Transformer Pooling & Deep Sets Pooling & Relational Network Pooling \\ \midrule
Ground Truth State  & \textbf{0.968}      & 0.880             & 0.870                      \\
SLATE               & \textbf{0.980}      & 0.513             & 0.150                      \\
SlotAttention-Large & \textbf{0.970}      & 0.650             & 0.410  \\ \bottomrule                   
\end{tabular}
    \caption{Property Comparison Task performance for Ground Truth State, SLATE, and SlotAttention-Large with additional pooling layers, Deep Sets \citep{deepset} and Relational Network \citep{rn}.}\label{appx:tab:additional_results_prop_comp_task}
\end{table}

\subsection{Does transformer pooling truly play a critical role in accurately extracting property-Level information?} \label{appx:trans_pooling_property_comparison}

To investigate this further, we conducted experiments on the Property Comparison Task, incorporating two additional pooling layers, Deep Sets \citep{deepset} and Relational Networks \citep{rn}. Our results in Table \ref{appx:tab:additional_results_prop_comp_task} reveal that the transformer pooling layer is vital for achieving optimal performance with OCRs, while the performances of Deep Sets and Relational Network pooling layers are considerably inferior. Interestingly, Ground Truth states exhibit strong performance with both Deep Sets and Relational Network pooling layers, attaining over 85\% accuracy. These findings imply that the transformer pooling layer plays a critical role in accurately extracting property-level information from OCRs. Moreover, our results indicate that SLATE, pre-trained through the transformer decoder, exhibits greater compatibility with the transformer pooling layer than SlotAttention-Large, as discussed in Appendix \ref{appx:trans_decoder_investigation}, since it demonstrates a larger performance gap between the transformer and other pooling layers.

\begin{figure}[h]
    \hspace{-8mm}
    \begin{center}
    \includegraphics[width=0.92\linewidth]{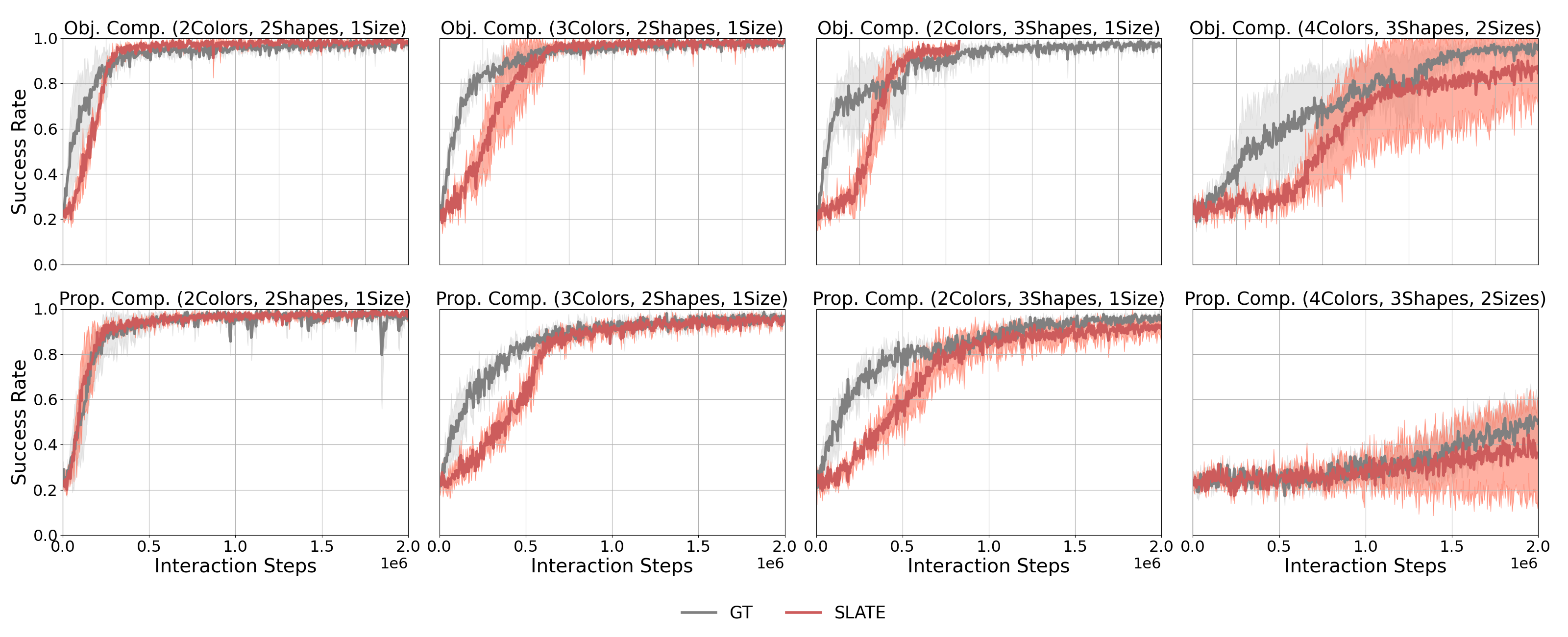}
    \vspace{-3mm}
    \end{center}
    \caption{Comparison of Success Rates as the difficulty level of the relational reasoning task increases.}
    \label{fig:stress_res}
    \vspace{-5mm}
\end{figure}

\subsection{How about OCR pre-training performance on more complicated relational-reasoning tasks?}

In this study, we investigate the performance of OCR pre-training on more complex relational-reasoning tasks. Previous study has demonstrated that OCR pre-training exhibits comparable performance to ground truth states on comparison tasks, as shown in Figures \ref{fig:overall_regime} and \ref{fig:overall}, and Table \ref{appx:tab:each_model_final_perf}. However, it is not clear if this performance extends to more complicated relational-reasoning tasks.

To address this question, we redesigned the comparison tasks to require more complex reasoning. Specifically, we extended the comparison tasks discussed in Appendix \ref{appx:benchmark_2d} to include three shapes or three colors, and evaluated the hardest condition in our benchmark, which consisted of one of four colors, three shapes, and two sizes.

The results, shown in Figure \ref{fig:stress_res}, indicate that as the task becomes harder, learning is slower, and all models failed to solve the Property Comparison task within 2 million steps when the object could have one of two sizes. However, OCR pre-training still exhibited comparable performance to ground truth states for all tasks, including the Property Comparison tasks.

Based on these findings, we hypothesize that the transformer pooling layer, when paired with a transformer decoder such as SLATE, is suitable for these comparison tasks and results in performance similar to ground truth states.

\begin{table}[h]
\centering
\begin{tabular}{ccc}
\toprule
      & \multicolumn{2}{c}{Success Rate}                                \\ \cmidrule(l){2-3}
Model & Object Comparison Task & Property Comparison Task \\ \midrule
SLATE & 0.963                  & 0.9                      \\ \bottomrule
\end{tabular}
    \caption{OCR pre-training performance for unseen objects in pre-training dataset.}\label{appx:tab:ood_ocr}
\end{table}

\begin{figure}[h]
    \hspace{-8mm}
    \begin{center}
    \includegraphics[width=0.45\linewidth]{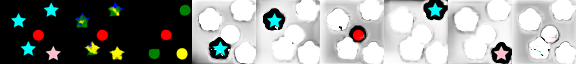}
    \hspace{3mm}
    \includegraphics[width=0.45\linewidth]{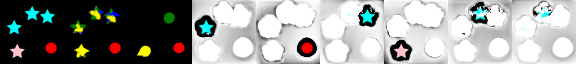}
    \vspace{-3mm}
    \end{center}
    \caption{Segmentation on the Property Comparison task with unseen objects in pre-training. From left, given image, reconstruction through DVAE, reconstruction through transformer decoder, and 6 slots.}
    \label{appx:fig:ood_ocr_samples}
    \vspace{-5mm}
\end{figure}

\subsection{OCR Pre-Training Performance for Unseen Objects in Pre-Training}

In this section, we investigate the possibility of OCR pre-training on an out-of-distribution environment to test its efficacy. Specifically, we evaluate the performance of the SLATE model on Object and Property Comparison tasks using color and shape sets [cyan, pink] and [star\_5, spoke\_4], which are unseen in the pre-training dataset.

As shown in Table \ref{appx:tab:ood_ocr}, our results indicate that the model achieves a success rate of over or approximately 90\% on the tasks, despite the objects being previously unseen. While this performance is lower than that achieved in the in-distribution environment, as demonstrated in Table \ref{appx:tab:each_model_final_perf}, it is still superior to the baselines. Our analysis of the segmentations depicted in Figure \ref{appx:fig:ood_ocr_samples} reveals that the model is capable of accurately segmenting objects, despite not being able to reconstruct the images perfectly. These findings suggest that OCR pre-training can be advantageous for out-of-distribution tasks if the model is capable of segmenting objects.

\begin{table}[h]
\centering
\begin{tabular}{cccccc} \toprule
                        &                     & \multicolumn{4}{c}{Models}                              \\\cmidrule(l){3-6}
                        &                     & SLATE  & Slot-Attention & Slot-Attention-Large & IODINE \\ \midrule
\multirow{4}{*}{FG-ARI} & Object Goal         & 0.910  & \textbf{0.928} & 0.913                & 0.918  \\
                        & Object Interaction  & 0.919  & \textbf{0.936} & 0.919                & 0.926  \\
                        & Object Comparison   & 0.912  & \textbf{0.929} & 0.914                & 0.920  \\
                        & Property Comparison & 0.911  & \textbf{0.930} & 0.916                & 0.922  \\ \midrule
\multirow{4}{*}{MSE}    & Object Goal         & 13.304 & \textbf{6.609} & 6.858                & 8.983  \\
                        & Object Interaction  & 93.919 & 47.368         & \textbf{46.133}      & 75.984 \\
                        & Object Comparison   & 10.185 & \textbf{5.033} & 6.053                & 8.168  \\
                        & Property Comparison & 9.276  & \textbf{4.732} & 5.904                & 8.431  \\ \bottomrule
\end{tabular}
    \caption{FG-ARI and MSE for 2D tasks}
    \label{appx:tab:ari_mse}
\end{table}

\begin{table}[h] \centering
\begin{tabular}{cccccc}
                                          &                     & \multicolumn{4}{c}{Models}                                              \\\cmidrule(l){3-6}
                                          &                     & SLATE          & Slot-Attention & Slot-Attention-Large & IODINE         \\\midrule
\multirow{4}{*}{$\text{R}^2$ for position}& Object Goal         & 0.858          & 0.512          & 0.613                & \textbf{0.997} \\
                                          & Object Interaction  & 0.875          & 0.741          & \textbf{0.899}       & 0.858          \\
                                          & Object Comparison   & 0.950          & 0.926          & \textbf{0.999}       & 0.995          \\
                                          & Property Comparison & 0.962          & \textbf{0.997} & 0.990                & 0.992          \\\midrule
\multirow{4}{*}{Color Accuracy}           & Object Goal         & \textbf{0.900} & 0.799          & 0.894                & 0.658          \\
                                          & Object Interaction  & \textbf{0.991} & 0.880          & 0.988                & 0.989          \\
                                          & Object Comparison   & 0.933          & 0.931          & 0.997                & \textbf{1.000} \\
                                          & Property Comparison & 0.979          & \textbf{1.000} & 0.999                & \textbf{1.000} \\\midrule
\multirow{4}{*}{Shape Accuracy}           & Object Goal         & 0.924          & 0.837          & 0.913                & \textbf{0.992} \\
                                          & Object Interaction  & \textbf{0.997} & 0.968          & \textbf{0.977}       & \textbf{0.997} \\
                                          & Object Comparison   & \textbf{0.977} & 0.968          & \textbf{0.997}       & \textbf{0.997} \\
                                          & Property Comparison & 0.933          & \textbf{1.000} & 0.997                & 0.998          \\\bottomrule
\end{tabular}
    \caption{Property prediction accuracy for 2D tasks. The accuracy for positions is calculated through $\text{R}^2$ score.}
    \label{appx:tab:prop_pred}
\end{table}

\begin{figure}[h]
    \centering
    \includegraphics[width=0.97\linewidth]{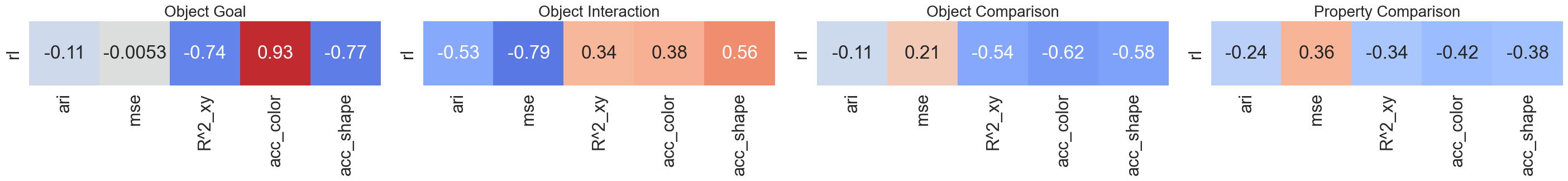}
    \caption{The correlation between performances on RL tasks and measurements such as FG-ARI, MSE and property prediction accuracies. Note that for MSE, positive correlation means that the model with smaller reconstruction error shows better RL performance.}
    \label{appx:fig:corr}
\end{figure}

\subsection{Do the standard metrics of evaluating OCR correlate with RL performance?} \label{appx:corr}

In this study, we aimed to investigate the correlation between standard OCR evaluation metrics, such as segmentation quality, reconstruction loss, and property prediction accuracy, and RL performance. To do this, we created task-specific datasets consisting of 50,000 training and 10,000 validation samples which were collected through a random policy on each task. The correlation is calculated using the foreground Adjusted Rand Index (FG-ARI) \citep{ari} and Mean Squared Error (MSE) for segmentation quality and reconstruction loss, respectively, and property prediction accuracy for SLATE, Slot-Attention, Slot-Attention-Large, and IODINE after training 50 epochs with 32 batch size. The results, presented in Figure \ref{appx:fig:corr}, indicate that while the FG-ARI showed a negative correlation with RL performance across all tasks, the correlation between MSE, property prediction accuracy, and RL performance was inconsistent, with positive correlation observed for some tasks and negative correlation for others. 

We further analyzed these results in Tables \ref{appx:tab:ari_mse} and \ref{appx:tab:prop_pred}. We observed that SLATE performed worse than the other models for FG-ARI and MSE in every task, but its RL performance was similar or better than the others. From these results, we hypothesize that when performance in segmentation or reconstruction is good enough, such as in the case of SLATE, the correlation with RL performance is weaker. We can find a similar trend for property prediction accuracy.

Interestingly, for the Object Goal task, Slot-Attention and Slot-Attention-Large performed slightly better than SLATE, despite their much lower position prediction accuracy. The reason for this is that exact position is not important for this task, as the task is considered solved if the agent goes near the target object. Conversely, IODINE's color prediction accuracy on the Object Goal task was much lower than the others, and it was correlated with its lower RL performance on this task. This can be attributed to color being an important factor in identifying the target object.

In conclusion, these results suggest that the standard OCR evaluation metrics may not be strongly correlated with RL performance, but sometimes they are correlated when the performance on the measurement is much worse than others, such as IODINE for the Object Goal task.

\section{Further Discussion} \label{appx:discussion}

In this study, we evaluated the effectiveness of OCR pre-training by using a benchmark that is limited to scenarios where a random policy can collect sufficient diverse observations. However, there are many scenarios where a random policy is not sufficient to pre-train the encoder. One potential future research direction is to investigate the use of auxiliary loss while training both the encoder and policy, as has been previously demonstrated in \citep{dreamer, worldmodel}. Additionally, it would be interesting to explore the use of OCR for non object-centric tasks, or to investigate the modulation between System 1 and System 2 modes \citep{thinking_fast_slow}.

\begin{table}[h] \centering
\begin{tabular}{ccccc} \toprule
                     &               \multicolumn{4}{c}{Tasks}                                      \\\cmidrule(l){2-5}
Models               & Object Goal    & Object Interaction & Object Comparison & Property Comparison \\ \midrule
GT                   & 0.974          & \textbf{0.795}     & 0.973             & 0.968               \\
CNN                  & \textbf{0.987} & 0.784              & 0.225             & 0.232               \\
CNNFeat              & 0.976          & 0.291              & 0.192             & 0.189               \\
MultiCNNs            & 0.974          & 0.490              & 0.537             & 0.567               \\
VAE                  & 0.674          & 0.340              & 0.386             & 0.387               \\
MAE-CLS              & 0.448          & 0.129              & 0.209             & 0.206               \\
SLATE-CNN            & 0.905          & 0.409              & 0.249             & 0.230               \\
MAE-Patch            & 0.901          & 0.049              & 0.216             & 0.219               \\
SLATE                & 0.985          & 0.787              & \textbf{0.979}    & \textbf{0.980}      \\
Slot-Attention       & 0.983          & 0.000              & 0.975             & 0.947               \\
Slot-Attention-Large & 0.986          & 0.047              & 0.967             & 0.970               \\
IODINE               & 0.916          & 0.003              & 0.698             & 0.624               \\ \bottomrule
\end{tabular}
    \caption{Success Rates after 2 million interaction steps. The performances are averaged from three random seeds.}
    \label{appx:tab:each_model_final_perf}
\end{table}

\begin{figure}[h]
    \centering
    \includegraphics[width=0.92\linewidth]{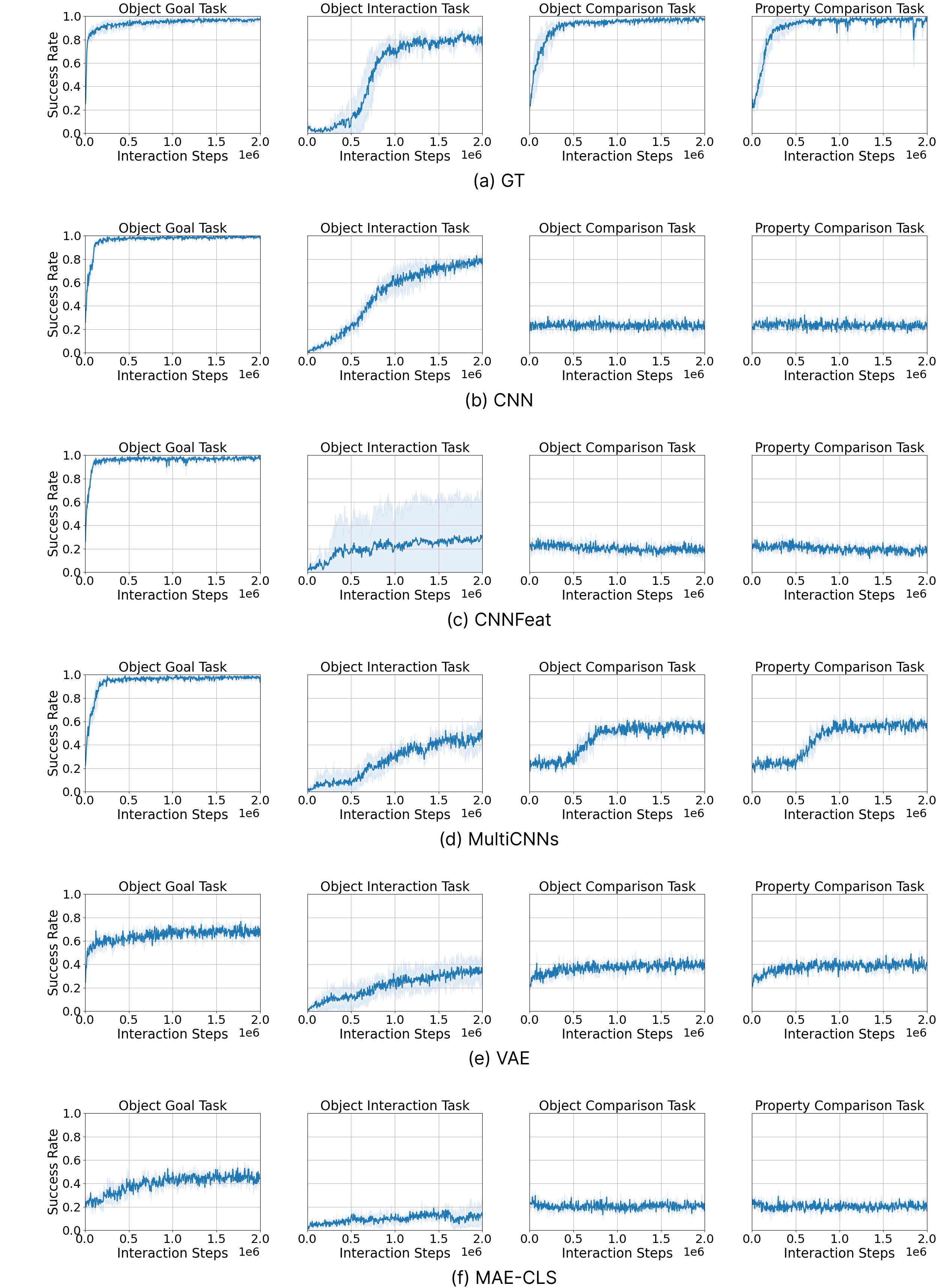}
    \caption{GT / CNN / CNNFeat / MultiCNNs / VAE / MAE-CLS performances for interaction steps}
\end{figure}

\begin{figure}[h]
    \centering
    \includegraphics[width=0.92\linewidth]{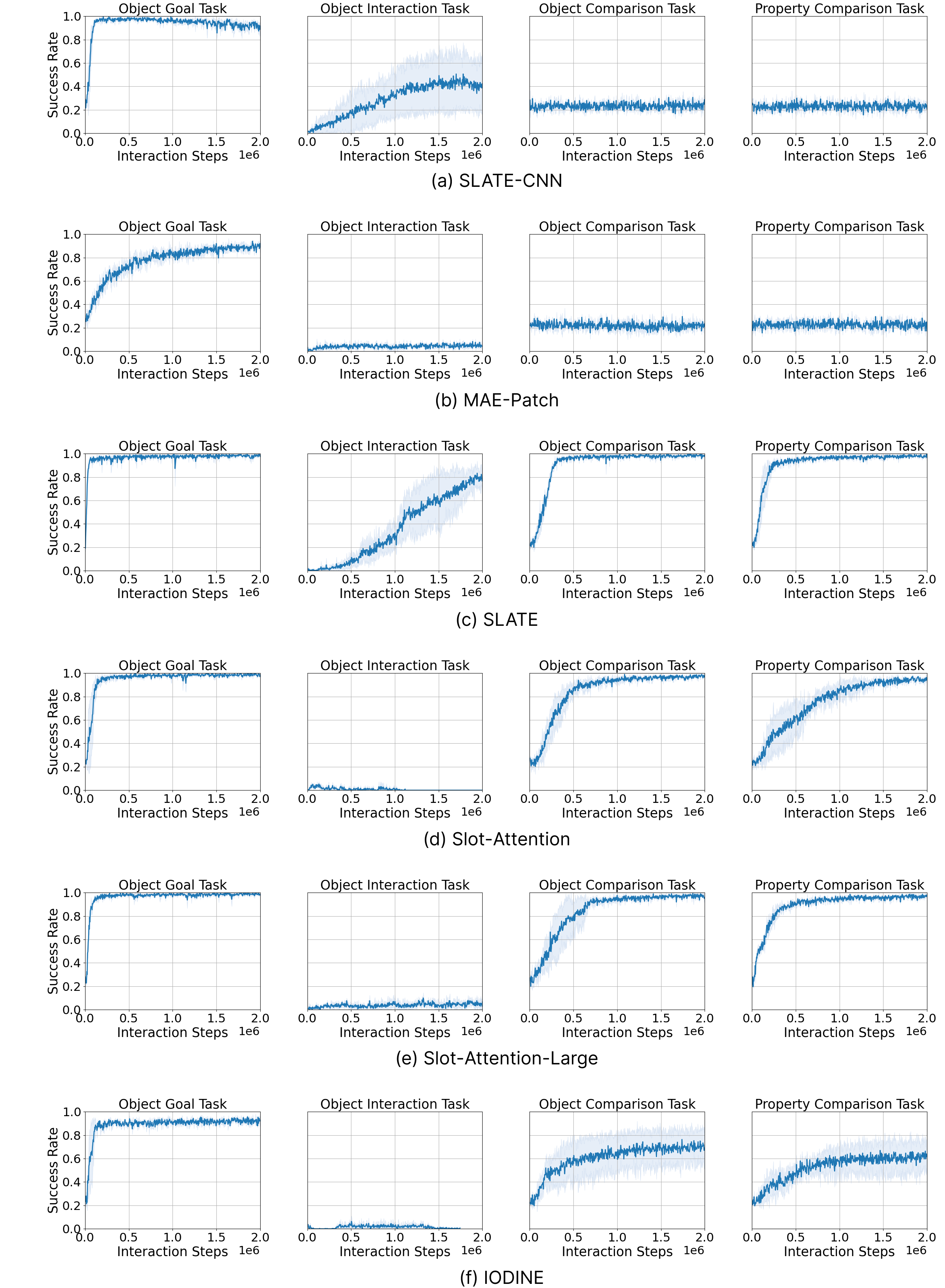}
    \caption{SLATE-CNN / MAE-Patch / SLATE / Slot-Attention / Slot-Attention-Large / IODINE performances for interaction steps}
\end{figure}

\begin{figure}[h]
    \centering
    \includegraphics[width=0.92\linewidth]{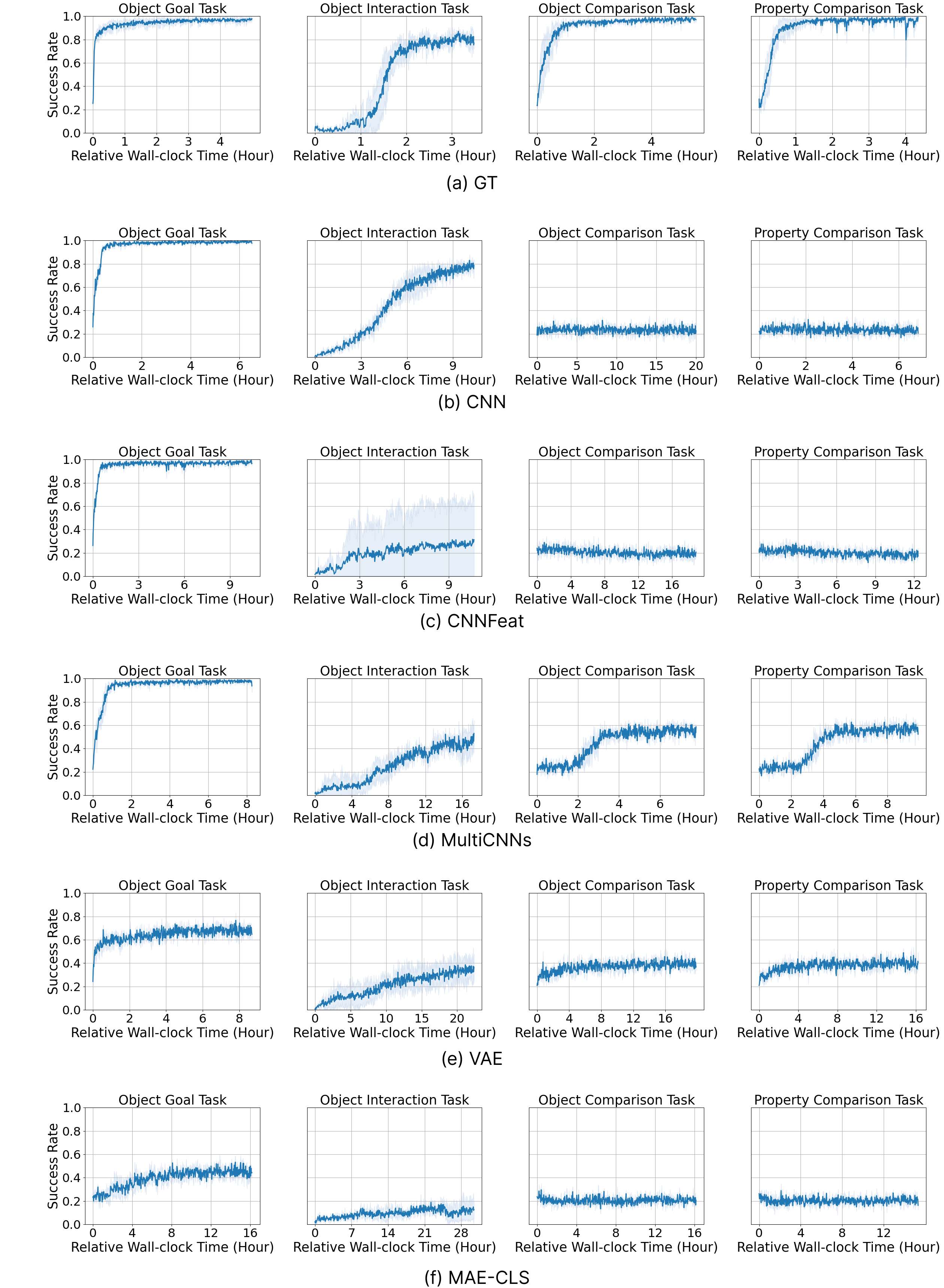}
    \caption{GT / CNN / CNNFeat / MultiCNNs / VAE / MAE-CLS performances for relative wall-clock time}
\end{figure}

\begin{table}[h]
\centering
\begin{tabular}{ccccccccc}
\toprule
                    & \multicolumn{4}{c}{Unseen \#Dists(Obj Goal)}              & \multicolumn{4}{c}{Unseen \#Dists(Obj Int.)} \\\cmidrule(l){2-5} \cmidrule(l){6-9}
Models              & 2            & 3(in)        & 4            & 5            & 0            & 1            & 2(in)        & 3        \\ \midrule
GT                  & 0.983        & 0.953        & 0.967        & 0.887        & nan          & nan          &\textbf{0.837}& nan      \\
CNN                 & 0.983        & 0.957        & 0.960        & 0.910        &\textbf{0.873}&\textbf{0.837}& 0.713        & 0.693    \\
CNNFeat             & 0.980        & 0.953        &\textbf{0.977}& 0.910        & 0.350        & 0.303        & 0.320        & 0.300    \\
MultiCNNs           & 0.990        & 0.950        & 0.960        & 0.923        & 0.527        & 0.550        & 0.443        & 0.450    \\
VAE                 & 0.680        & 0.623        & 0.610        & 0.577        & 0.460        & 0.343        & 0.327        & 0.353    \\
MAE-CLS             & 0.407        & 0.427        & 0.403        & 0.357        & 0.043        & 0.117        & 0.147        & 0.093    \\
SLATE-CNN           & 0.987        &\textbf{0.967}& 0.973        &\textbf{0.930}& 0.523        & 0.427        & 0.413        & 0.423    \\
MAE-Patch           & 0.983        & 0.950        & 0.930        & 0.857        & 0.073        & 0.043        & 0.047        & 0.067    \\
SLATE               & 0.990        & 0.937        & 0.963        & 0.907        & 0.790        & 0.813        & 0.747        & \textbf{0.700}\\
SlotAttention       &\textbf{0.997}& 0.960        & 0.960        & 0.920        & 0.063        & 0.033        & 0.033        & 0.057    \\
SlotAttention-Large & 0.987        &\textbf{0.967}& 0.970        & 0.907        & 0.063        & 0.040        & 0.057        & 0.057    \\
IODINE              & 0.960        & 0.890        & 0.887        & 0.827        & 0.030        & 0.023        & 0.026        & 0.026    \\ \bottomrule
\end{tabular}
    \caption{The performances for unseen number of objects on Object Goal and Interaction tasks. Note that GT model cannot be validated for unseen number cases on Object Interaction task, because the used architecture cannot support that as described in \textbf{Question 3}.}
    \label{appx:tab:unseen_num1}
\end{table}

\begin{table}[h]
\centering
\begin{tabular}{ccccccccc}
\toprule
                    & \multicolumn{4}{c}{Unseen \#Objs(Obj Comp.)}              & \multicolumn{4}{c}{Unseen \#Objs(Prop Comp.)} \\\cmidrule(l){2-5} \cmidrule(l){6-9}
Models              & 3            & 4(in)        & 5            & 6            & 3            & 4(in)        & 5            & 6         \\ \midrule
GT                  &\textbf{0.967}& 0.933        & 0.943        & 0.870        &\textbf{0.977}& 0.943        & 0.940        & 0.903     \\
CNN                 & 0.257        & 0.207        & 0.240        & 0.203        & 0.240        & 0.263        & 0.210        & 0.177     \\
CNNFeat             & 0.287        & 0.213        & 0.203        & 0.163        & 0.247        & 0.227        & 0.213        & 0.147     \\
MultiCNNs           & 0.527        & 0.500        & 0.533        & 0.503        & 0.540        & 0.587        & 0.537        & 0.457     \\
VAE                 & 0.370        & 0.383        & 0.400        & 0.323        & 0.330        & 0.400        & 0.400        & 0.340     \\
MAE-CLS             & 0.210        & 0.197        & 0.180        & 0.127        & 0.187        & 0.140        & 0.173        & 0.177     \\
SLATE-CNN           & 0.257        & 0.200        & 0.200        & 0.160        & 0.240        & 0.187        & 0.203        & 0.200     \\
MAE-Patch           & 0.263        & 0.223        & 0.197        & 0.217        & 0.257        & 0.237        & 0.197        & 0.157     \\
SLATE               & 0.917        &\textbf{0.950}&\textbf{0.973}&\textbf{0.923}& 0.963        & 0.927        & 0.933        &\textbf{0.913}     \\
SlotAttention       & 0.903        & 0.937        & 0.943        & 0.880        & 0.930        & 0.920        & 0.883        & 0.827     \\
SlotAttention-Large & 0.913        & 0.930        & 0.943        & 0.853        & 0.967        &\textbf{0.957}&\textbf{0.950}& 0.897     \\
IODINE              & 0.703        & 0.700        & 0.690        & 0.627        & 0.593        & 0.640        & 0.613        & 0.523     \\ \bottomrule
\end{tabular}
    \caption{The performances for unseen number of objects on Object and Property Comparison tasks.}
    \label{appx:tab:unseen_num2}
\end{table}

\begin{table}[h]
\centering
\begin{tabular}{ccccccccc}
\toprule
                    & \multicolumn{4}{c}{Unseen Colors(Obj Goal)}               & \multicolumn{4}{c}{Unseen Colors(Obj Int.)} \\\cmidrule(l){2-5} \cmidrule(l){6-9}
Models              & 0(in)        & 1            & 2            & 3            & 0(in)     & 1         & 2        & 3        \\ \midrule
GT                  & 0.953        & 0.937        & 0.900        & 0.930        &\textbf{0.837}&\textbf{0.800}&\textbf{0.740}&\textbf{0.747}    \\
CNN                 & 0.957        & 0.960        & 0.960        & 0.960        & 0.713        & 0.693        & 0.697        & 0.683    \\
CNNFeat             & 0.953        & 0.953        & 0.953        & 0.950        & 0.320        & 0.307        & 0.310        & 0.300    \\
MultiCNNs           & 0.950        & 0.960        & 0.950        & 0.940        & 0.443        & 0.447        & 0.463        & 0.433    \\
VAE                 & 0.623        & 0.597        & 0.670        & 0.397        & 0.327        & 0.287        & 0.323        & 0.227    \\
MAE-CLS             & 0.427        & 0.380        & 0.410        & 0.310        & 0.147        & 0.097        & 0.103        & 0.070    \\
SLATE-CNN           &\textbf{0.967}&\textbf{0.980}&\textbf{0.970}& 0.847        & 0.413        & 0.417        & 0.407        & 0.250    \\
MAE-Patch           & 0.950        & 0.950        & 0.925        & 0.897        & 0.047        & 0.070        & 0.037        & 0.043    \\
SLATE               & 0.937        & 0.953        & 0.943        & 0.893        & 0.747        & 0.750        & 0.720        & 0.713    \\
SlotAttention       & 0.960        & 0.947        & 0.963        &\textbf{0.963}& 0.033        & 0.017        & 0.047        & 0.047    \\
SlotAttention-Large &\textbf{0.967}& 0.950        & 0.947        & 0.940        & 0.057        & 0.047        & 0.040        & 0.027    \\
IODINE              & 0.890        & 0.907        & 0.893        & 0.893        & 0.027        & 0.033        & 0.013        & 0.027    \\ \bottomrule
\end{tabular}
    \caption{The performances for unseen colors on Object Goal and Interaction tasks.}
    \label{appx:tab:unseen_color1}
\end{table}

\begin{table}[h]
\centering
\begin{tabular}{ccccccc}
\toprule
                    & \multicolumn{3}{c}{Unseen Colors(Obj Comp.)} & \multicolumn{3}{c}{Unseen Colors(Prop Comp.)} \\ \cmidrule(l){2-4} \cmidrule(l){5-7}
Models              & 0(in)         & 1             & 2            & 0(in)         & 1             & 2             \\ \midrule
GT                  & 0.933         & 0.480         & 0.183        & 0.943         &\textbf{0.663} & 0.293         \\
CNN                 & 0.207         & 0.213         & 0.203        & 0.263         & 0.157         & 0.220         \\
CNNFeat             & 0.213         & 0.233         & 0.203        & 0.227         & 0.227         & 0.200         \\
MultiCNNs           & 0.500         & 0.273         & 0.267        & 0.587         & 0.280         & 0.247         \\
VAE                 & 0.383         & 0.287         & 0.230        & 0.400         & 0.333         & 0.297         \\
MAE-CLS             & 0.197         & 0.197         & 0.163        & 0.140         & 0.163         & 0.177         \\
SLATE-CNN           & 0.200         & 0.193         & 0.157        & 0.187         & 0.230         & 0.210         \\
MAE-Patch           & 0.223         & 0.223         & 0.180        & 0.237         & 0.227         & 0.240         \\
SLATE               &\textbf{0.950} & 0.637         & 0.647        & 0.927         & 0.593         &\textbf{0.590} \\
SlotAttention       & 0.937         & 0.600         & 0.377        & 0.920         & 0.513         & 0.357         \\
SlotAttention-Large & 0.930         &\textbf{0.923} &\textbf{0.650}&\textbf{0.957} & 0.453         & 0.443         \\
IODINE              & 0.700         & 0.617         & 0.280        & 0.640         & 0.520         & 0.320         \\ \bottomrule
\end{tabular}
    \caption{The performances for unseen colors on Object and Property Comparison tasks.}
    \label{appx:tab:unseen_color2}
\end{table}


\begin{table}[h]
\centering
    \begin{tabular}{@{}lccc@{}}
    \toprule
    \multicolumn{2}{c}{Configurations}  & SLATE \\ \midrule
    \multirow{10}{6em}{Learning} &Temp. Cooldown   &  1.0 to 0.1  \\ 
    &Temp. Cooldown Steps & 30000 \\ 
    &LR for DVAE & 0.0003 \\ 
    &LR for CNN Encoder & 0.0001 \\
    &LR for Transformer Decoder & 0.0003 \\
    &LR Warm Up Steps & 30000 \\
    &LR Half Time & 250000 \\
    &Dropout & 0.1 \\
    &Clip & 0.05 \\
    &Batch Size & 24 \\ \midrule
    \multirow{1}{6em}{DVAE } &vocabulary size &  4096  \\ \midrule
    \multirow{1}{6em}{CNN Encoder} &Hidden Size &  64  \\ \midrule
    \multirow{6}{6em}{Slot Attention} &Slots &  -   \\ 
    &Iterations   &  3   \\ 
    &Slot Heads & 1  \\ 
    &Slot Dim.   &  192   \\ 
    &MLP Hidden Dim. & 192  \\ 
    &Pos Channels & 4 \\\midrule
    \multirow{3}{6em}{Transformer Decoder} &Layers &  4  \\ 
    &Heads   &  4  \\ 
    &Hidden Dim. & 192 \\ \bottomrule
    \end{tabular}
    \caption{Hyperparameters for SLATE}
    \label{appx:tab:slate}
\end{table}

\begin{table}[h]
\centering
    \begin{tabular}{@{}lcccc@{}}
    \toprule
    \multicolumn{2}{c}{Configurations}& Slot-Attention & Slot-Attention (Large) \\ \midrule
    \multirow{5}{6em}{Learning} &LR & 0.0001 & 0.0001 \\
    &LR Warm Up Steps & 30000 & 30000\\
    &LR Half Time & 250000 & 250000\\
    &Clip & 0.05 & 0.05\\
    &Batch Size & 24 & 24\\ \midrule
    \multirow{1}{6em}{CNN Encoder} &Hidden Size &  64 & 64 \\ \midrule
    \multirow{6}{6em}{Slot Attention} &Slots &  -  & - \\ 
    &Iterations   &   7 & \textbf{3} or \textbf{7}  \\ 
    &Slot Heads & 1 &  1 \\ 
    &Slot Dim.    & 64 & \textbf{192}  \\ 
    &MLP Hidden Dim.  & 128 & \textbf{192} \\
    &Pos Channels & 4 & 4\\ \bottomrule
    \end{tabular}
    \caption{Hyperparameters for Slot-Attention and Slot-Attention (Large)}
    \label{appx:tab:sa}
\end{table}



\end{document}